# AutoFPDesigner: Automated Flight Procedure Design Based on Multi-Agent Large Language Model

Longtao Zhu [a], Hongyu Yang[a,*], Ge Song[c], Xin Ma[c], Yanxin Zhang[b], Yulong Ji [b,*],

[a]*College of Computer Science, Sichuan University, Chengdu 610065, China*
[b]*School of Aeronautics and Astronautics, Sichuan University, Chengdu 610065, China*
[c]*College of air traffic management, Civil Aviation Flight University of China, Chengdu 618307, China*

**Abstract**

Current flight procedure design methods heavily rely on human-led design process, which is not only low automation but also suffer from complex algorithm modelling and poor generalization. To address these challenges, this paper proposes an agent-driven flight procedure design method based on large language model, named AutoFPDesigner, which utilizes multi-agent collaboration to complete procedure design. The method enables end-to-end automated design of performance-based navigation (PBN) procedures. In this process, the user input the design requirements in natural language, AutoFPDesigner models the flight procedure design by loading the design specifications and utilizing tool libraries complete the design. AutoFPDesigner allows users to oversee and seamlessly participate in the design process. Experimental results show that AutoFPDesigner ensures nearly 100% safety in the designed flight procedures and achieves 75% task completion rate, with good adaptability across different design tasks. AutoFPDesigner introduces a new paradigm for flight procedure design and represents a key step towards the automation of this process.[1*]

## 1. Introduction

The design of Instrument Flight Procedures (IFP) is currently a manual process led by humans, which strictly adheres to design specification documents[1].With the introduction and widespread adoption of Performance-Based Navigation (PBN) flight procedures, the process of designing PBN flight procedures according to these specifications has become increasingly complex. Consequently, the human-led design process has become more time-consuming[1].

Since last year, the rapid advancement of large language models has been impressive, particularly with the emergence of generative models like ChatGPT[3]. These models have achieved significant success in text classification, question answering, and data processing[4], and have also shown considerable potential in code debugging, game design, and architecture[5-7]. The latter three tasks fundamentally involve operating based on standard specifications, which is similar to flight procedure design. Therefore, the human-like text comprehension and planning capabilities of large language models present unprecedented opportunities for the automation of flight procedure design. Rather than merely assisting human designers, our work takes a further step by enabling large language models to lead the flight procedure design based on design specifications, with human designers supervising and seamlessly participating in the design process.

Compared to using a single large model as an agent for reasoning and planning, multi-agent systems based on Large Language Models (LLMs) hold greater potential for solving complex problems[8]. Multi-agent systems, akin to human collaboration in tackling intricate issues, involve multiple autonomous agents engaging in discussion, planning, and decision-making. Recent studies have explored the use of multi-agent systems to address complex problems in areas such as software development[9], social simulation[10], and game simulation[11]. Flight procedure design is a complex problem that involves adhering to numerous specifications, performing specialized calculations, and extracting critical navigation data. By employing a multi-agent approach based on LLMs, we can achieve agent-driven flight procedure design. In our method, individual agents simulate specific aspects of the flight procedure design, and then collaborate to discuss and make decisions based on human design requirements, ultimately realizing automated flight procedure design.

Taking a step towards more intelligent automation of flight procedure design, we present AutoFPDesigner, an agent empowered by multi-agent large language models. The core motivation of our approach is the ability of large language models to understand complex design specifications described in natural language at a level comparable to that of professional designers. By leveraging LLM agents as interactive schedulers between design specifications and specialized tools, our approach fundamentally changes the flight procedure design paradigm.

We summarize our contributions as follows:

(a) We propose AutoFPDesigner, an LLM-driven agent that fundamentally transforms the primary entity driving flight procedure design, providing a powerful and flexible paradigm for the automation and intelligent design of flight procedures.

(b) We have created a json-format navigation dataset for multiple airports, sourced from AIP23. This dataset includes information on airports, runways, flight procedures, waypoints, and obstacles. Compared to the original data format, this database format is more suitable for research related to flight procedure design.

(c) AutoFPDesigner integrates a multifunctional tool library, multiple agents with specialized knowledge, and essential navigation data, all orchestrated by LLM to achieve professional and accurate agent-driven automated flight procedure design.

(d) We tested the performance of AutoFPDesigner at Shuangliu and Jiangbei Airport. The experimental results show that the departure flight procedures designed by AutoFPDesigner can achieve almost 100 % safety, meet the special constraints of the departure procedures, and have a task completion rate of about 75%.

## 2. Relate work

### 2.1. Automatic flight procedure design

Flight procedure design can be modeled as a path planning problem subject to various constraints. Many researchers have adopted heuristic or machine learning methods to design flight procedures. The methods mentioned in reference[12-15] are capable of generating flight paths that meet the design specifications. However, they require redesigning the model for each airport's flight procedures, a process that depends heavily on specialized algorithm researchers. In contrast, AutoFPDesigner models flight procedure design through natural language, making it more accessible and understandable to professionals who are not experts in heuristic or machine learning algorithms. Additionally, when switching airports, only the navigation data within AutoFPDesigner needs to be updated.

Additionally, flight procedure designers often use commercial software such as AutoCAD[16]and IDS for designing flight procedures. These commercial tools provide professional support for route planning and measurement, with the design process being primarily led by the designers themselves. This paradigm relies on designers who are professionally trained and continuously updated with the design specifications, making the performance of the designed flight procedures dependent on the expertise of the designers. In contrast, AutoFPDesigner introduces an LLM-based agent to lead the flight procedure design process, fundamentally changing the traditional design paradigm. This agent-driven approach leverages large language models to automate the design process, reducing the dependence on specialized human expertise.

### 2.2. Multi-agent in LLMs

Generative large language models were initially proposed by OpenAI[17] Models like GPT-3.5 and GPT-4 have demonstrated impressive capabilities in conversation and reasoning. To better harness the power of LLMs, researchers have introduced methods such as Reinforcement Learning from Human Feedback (RLHF), Chain-of-Thought (CoT), and ReAct[18-20] to enhance the reasoning and decision-making abilities of LLMs, thereby improving their general capabilities. Through fine-tuning, LLMs can be adapted to excel in specific professional domains. These advancements have led to significant success in language understanding and generation, showing immense potential in addressing complex professional problems.

For complex professional problems, previous works have utilized LLMs to call tools and execute sequential actions, as seen in models like Auto-GPT and AgentGPT[21]. AutoGen and MetaAgents[22-24] have proposed incorporating multiple agents to tackle complex issues, with the core idea being that multi-agent systems can emulate human societal collaboration, the foundation of collective human intelligence. Collaboration allows for breaking down complex problems into simpler sub-tasks, and LLMs have already shown exceptional capability in solving individual tasks. Moreover, interactions between different agents can foster divergent thinking, enhance factual accuracy[25], reasoning[26] and provide validation[27].

In the context of flight procedure design, our insight is that it is a complex process comprising multiple sub-tasks such as specification understanding, data extraction, tool invocation, and route design. Leveraging the power of LLMs in flight procedure design remains an open challenge. In this paper, we address this challenge by applying multi-agent LLMs[28-32] to language modeling of the flight procedure design process. To the best of our knowledge, this is the first work to apply LLMs to flight procedure design.

# 3. AutoFPDesigner

In this section, we will provide a comprehensive overview of AutoFPDesigner, an intelligent flight procedure design agent based on a large language model. Section 3.1 will present an introduction to the overall framework. Subsequently, we will delve into the three key components of our approach: Rules & Memory (Section 3.2), Tool Library (Section 3.3), and the LLM Agents (Section 3.4).

## 3.1. Overall Architecture

Traditionally, flight procedure design is performed by experts using various specialized tools to meet specific requirements. The overall framework of AutoFPDesigner is shown in Fig. 1 with the key distinction being that an agent, rather than a human designer, drives the design process. The Rules & Experience Memory endows AutoFPDesigner with the specialized knowledge and design experience necessary for flight procedure design, while the Tool Library provides access to the professional tools required for the task. The LLM Agents refine the work of human designers by breaking down their tasks into specific components such as understanding user requirements, overall route planning, detailed waypoint planning, latitude and longitude calculations, and visual display. Each component is handled by a dedicated agent. This multi-agent workflow approach is designed to more effectively accomplish design tasks compared to a single agent handling all tasks, as it lowers the baseline requirements for the capabilities of individual language models.

## 3.2. Rules&Experience Memory

Flight procedure design experts rely on their continually updated professional knowledge and experience to design flight procedures. This expertise fundamentally stems from the flight procedure design standards outlined in ICAO Doc 8168, which are represented in textual form. Large language models (LLMs) naturally excel in understanding textual language, making them inherently suited to grasp these design standards. In contrast, modeling such standards within other design methodologies can be challenging. Specifically, text-based design standards are optimally modeled by LLMs without additional processing, whereas other design approaches require modeling these standards through design objective functions. For each flight procedure design requirement, we leverage existing design experience to aid in decision-making. Expert Knowledge comprises two key components: Rules and Experience Memory.

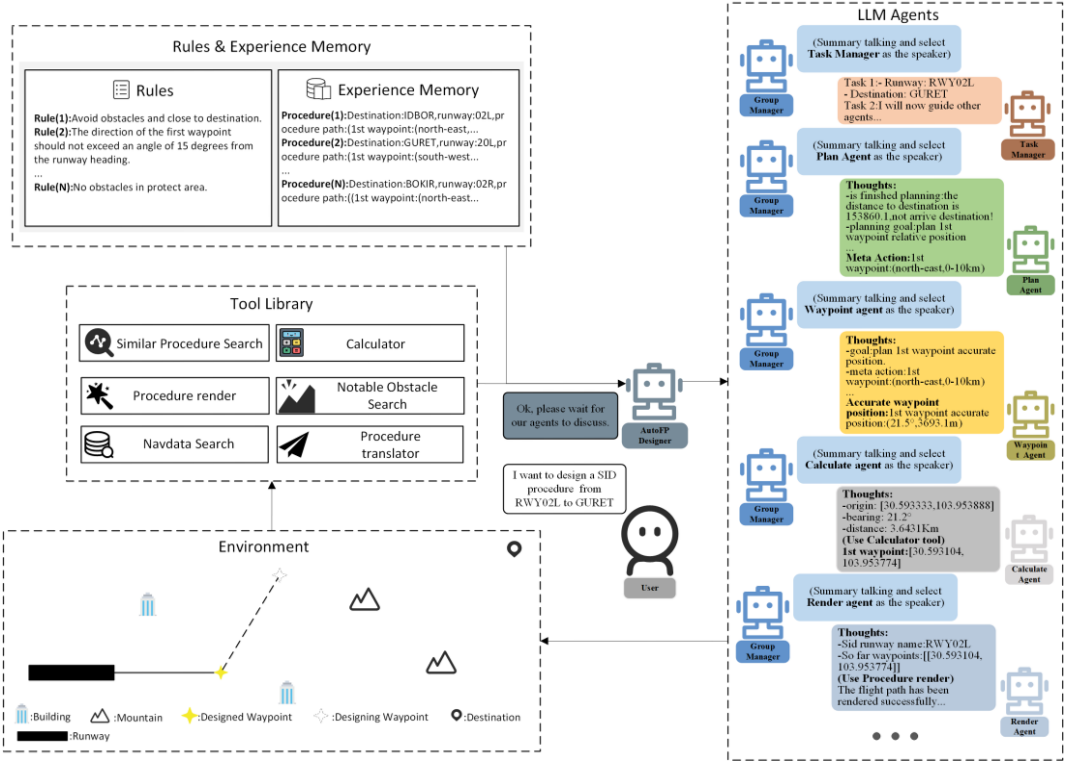

Fig. 1 The overall architecture of AutoFPDesigner.

**Rules:** The fundamental principles of flight procedure design, such as avoiding notable obstacles and setting initial trajectory constraints, are encapsulated within the system. Notably, the Rules are entirely text-based, allowing users to configure them for different types of flight procedures. This enables the customization and enhancement of AutoFPDesigner's capability to design various types of flight procedures with greater specialization and precision.

**Experience Memory:** The Experience Memory includes a selection of existing flight procedures for the airport, with each entry consisting of the design purpose, obstacle information, and subsequent design decisions. By retrieving the most similar experiences and referencing their design decisions, AutoFPDesigner enhances its alignment with waypoint reuse and human-like design processes. The data format for these entries is structured as [Destination, Runway, Procedure Path].

The Rules & Experience Memory endows AutoFPDesigner with both professional knowledge of flight procedure design and the design experience of past procedures. The retrieved most similar procedures, along with the expert knowledge, collectively form one of the inference inputs for AutoFPDesigner. Modeling flight procedure design expertise as text increases the extensibility and compatibility of AutoFPDesigner.

*3.3. Tools Library*

The motivation for using the tool library is to expand agent's capabilities beyond verbal communication. The way AutoFPDesiger calls the tools is to come is to implement the navigation data search, obstacle evaluation and design decisions required for the flight procedure design, and all of the tools have returned results in the form of text. The tool library consists of six tools: Navadata Search, Similar Procedure Search, Notable Obsacle Search, calculator, Procedure Render and Procedure Transfer. it is worth noting that the tool library is supported by the extensions to meet different flight procedure design needs.

**Navadata Search:** This tool assist AutoFPDesigner in obtaining accurate initial environmental information based on user design requirements, such as runway coordinates, headings, and procedure termination point coordinates.

**Similar Procedure Search:** In the Experience Memory, the most similar flight procedures to the current design are returned. These similar flight procedures include two types: the most similar procedure departing from the same runway and the most similar procedure terminating at the same endpoint. For procedures departing from the same runway, similarity is calculated by:

$$S_1 = \frac{1}{\Delta d(Ewpt_{p_1}, Ewpt_{p_{design}})}, p_1 \in MatchP_1 \qquad (1)$$

where $S_1$ represents the similarity score, $MatchP_1$ denotes the set of procedures within the Experience Memory that share the same departure runway as the procedure being designed, $Ewpt_p$ represents the termination waypoint of procedure, $\Delta d$ denotes the distance between two waypoints, the highest similarity score being selected as the return.

For procedures terminating at the same endpoint, similarity is calculated by:

$$S_2 = \frac{1}{\Delta\theta(Heading_{p_2}, Heaidng_{p_{design}})}, p_2 \in MatchP_2 \qquad (2)$$

where $S_2$ represents the similarity score, $MatchP_2$ denotes the set of procedures within the Experience Memory that share the same termination waypoint as the procedure being designed, $Heading_p$ represents the departure runway heading of the procedure, $\Delta\theta$ the angular difference between two runway headings, the highest similarity score being selected as the return.

**Notable Obstacle Search:** Retrieve threatening obstacles within a 50 km radius. These obstacles pose a threat when designing flight procedures in their direction, as they extend beyond the obstacle evaluation surfaces.

**Calculator:** Based on AutoFPDesigner's decision output, the precise coordinates of the next waypoint are calculated. The decision output format is [azimuth, distance]. The Calculator uses the current waypoint coordinates along with the [azimuth, distance] to compute the latitude and longitude coordinates of the next waypoint.

**Procedure Render:** During the design process, visualize the flight procedures designed by AutoFPDesigner, including the flight route, protection zones, and obstacles. Based on the visualization results, users can choose whether to provide design feedback to AutoFPDesigner, this tool is invoked when human users are involved in the design process

**Procedure transfer:** Convert the completed flight procedures into a specific format, in this paper converted to txt, for safety assessment.

*3.4. Multi-Agent LLMs*

For the complex task of flight procedure design, to reduce the occurrence of hallucinations in large models, this paper adopts a multi-agent collaborative approach to tackle this challenge. The advantage of this method is that it lowers the minimum performance threshold required for the agents and reduces the frequency of hallucinations. In this paper, seven agents are configured with the following functions:

**Group Manager:** Selects the next speaking agent based on user requirements and historical dialogue.

**Task Agent:** Responsible for refining user input, such as extracting the runway number and the type of procedure to be designed as specified by the user.

**Plan Agent:** Utilizes Navadata Search, Similar Procedure Search, Notable Obstacle Search, and user requirements to perform a preliminary planning of the flight procedure, outputting the approximate azimuth and distance to the next waypoint.

**Waypoint Agent:** Based on the preliminary

planning from the Plan Agent and using the Obstacle Search tool, provides the precise location of the next waypoint, specifically the azimuth and distance to the next waypoint.

**Calculate Agent:** Using the Calculator tool, computes the exact latitude and longitude of the next waypoint based on the Waypoint Agent's input and the current waypoint's coordinates.

**Render Agent:** Utilize the Procedure Render and Procedure transfer tools to visualize the designed waypoints and corresponding protection zones, allowing users to intuitively monitor the design process. Additionally, store the completed procedures in TXT format.

**FixWaypoint Agent:** This agent is used when humans are involved in the interaction. It is responsible for interpreting human feedback and making modifications to the flight procedure waypoints based on that understanding.

These agents prompt can be found in Appendix A1.

### 3.5. Fine-tuning&Chain-of-Thought

In addition to employing a multi-agent approach to mitigate hallucinations in AutoFPDesigner, we also utilize fine-tuning and chain-of-thought techniques to enhance the reasoning capabilities of AutoFPDesigner. We applied these techniques to the Plan Agent and Waypoint Agent to ensure their outputs follow specific reasoning patterns and formats. These two agents are central to the autonomous design of flight procedures. Additionally, we employed reasoning chains to enhance these two agent inferential capabilities.

**Fine-tuning:** Fine-tuning is a technique used to enhance the specific capabilities of a large model and has been widely applied in various domains. In this work, we employed fine-tuning for the Plan Agent and Waypoint Agent, with the number of fine-tuning data points shown in Fig. 2 For the fine-tuning of the Plan Agent, the inputs include the flight procedure name, similar procedure plans, obstacle information, and distance to the endpoint. The output is the approximate plan for the next waypoint. For the fine-tuning of the Waypoint Agent, the inputs consist of the approximate plan of the previous waypoint, obstacle information, and distance to the endpoint. The output is the detailed planning information for the next waypoint.

**Chain-of-Thought:** Chain-of-thought is a widely-used technique to enhance the reasoning capabilities of large models. The core idea is to have the model output its thought process step-by-step before providing the final result, thereby improving its reasoning ability. This technique has been implemented in both the Plan Agent and Waypoint Agent in our study.

## 4. Experiments

In this section, we demonstrate the effectiveness, compatibility and other features of AutoFPDesigner through extensive experiments on the design of departure flight procedures at Shuangliu(ZUUU) and Jiangbei(ZUCK) airports. We first describe the experimental setup and evaluation metrics in Section 4.1. Next we investigate AutoFPDesigner's few-shot learning capabilities (Section 4.2), compatibility (Section 4.3), stability (Section 4.4) and interpretability (Section 4.5) as well as ablation experiments (Section 4.6) to evaluate the effectiveness of the different components in AutoFPDesigner, and finally demonstrate the effectiveness of AutoFPDesigner in the human-involved(Section 4.7).

### 4.1. Experimental Setup

**Dataset.** To fine-tune the Plan Agent and Waypoint Agent, we have developed a PBN flight procedure dataset in JSON format, derived from AIP23 data. The dataset currently includes approximately 40 departure procedures from Shuangliu and Jiangbei airports, encompassing over 300 flight segments. It also contains information on airport obstacles and runway details. The dataset is available at the following GitHub repository: https://github.com/Zhulongtao6/FPDataSet.git. To the best of our knowledge, this is the first open-source flight procedure dataset available.

**Implementation details.** In our experiments we used AutoFPDesigner to design straight-line departures flight procedure, and the prompts for each agent in AutoFPDesigner are centred around this type flight procedure.

**Evaluation metrics.** We propose evaluation metrics for AutoFPDesigner based on three aspects: safety, type, and efficiency. Safety, the foremost consideration in flight procedure design, is assessed by evaluating the presence of obstacles within protected zones. Flight procedures encompass various types, and in addition to meeting general standards, each type must adhere to specific constraints. Efficiency measures whether AutoFPDesigner can complete the design within a limited number of interactions. The proposed metrics corresponding to these criteria are Flight Procedure Safety (FPS), Special Constraints Compliance (SCC), and Mission Completion Rate (MCR).

*Flight Procedure Safety(FPS):* The safety of the flight procedure is achieved by the absence of obstacles in the protected area, as shown in the Fig. 3 the protected area is divided into two parts, horizontal and vertical. In the horizontal part, the protection zone is divided into two parts, the main area and the sub-area, and the width of the protection area is determined by the navigation specification of the flight procedure[10], for the straight-line departure flight procedure generally adopts the navigation specification of RNP1, and its horizontal width of the protection zone is 4360m. In the vertical part, the minimum height of the protection area is determined by the height of the highest obstacle in the segment where it is located, and the height of the highest obstacle in the minimum height needs to be increased on the basis of the height of the highest obstacle, which is called minimum obstacle clearance (MOC).

******System prompt:******

'n**RNP SID Flight Procedure Path Top Designer** nRole:You are the SID RNP Flight Procedure Path Top Designer, plan relative position and distance of next waypoint in sequence based on offered information. Avoid obstacles and close to destination. 'n nContext:n- Coordinates: X-axis is the east, Y-axis is the north, When planning the next waypoint you are at the last waypoint as the origin(0,0), heading:'0 is north 'n n-first waypoint restriction: the direction of the first waypoint should not exceed an angle of 15 degrees from the runway heading.'n -similar procedure: the similar sid runway procedure and the similar destination procedure can offer advice for your plan.'n-Objective:plan next waypoint in sequence based on offered information.'n'nInputs:n1...

******User Input: ******

the planning sid procedure name:BOKIR-9W nsid runway:RWY02L, nrunway heading:21.8°,nrunway name:02L ndestination:(name:BOKIR,relative position:358.29°,distance:59083.2m) nsimilar sid procedure path from same sid runway: sid runway:02L,destination:IDBOR,procedure path:(1st waypoint,dis to destination:162529.8,meta action (0~45°,0~10km))...

******Assistant Output: ******

Thoughts:
-is finished planning:the distance to destination is 59083.2,not arrive destination!
-planning goal:plan 7th waypoint relative position from RWY02L to BOKIR.
-notable obstacles:(name:LL,relative position:270.71°,distance:48965.5m) navoid these obstacles
-similar sid procedure path from same sid runway: sid runway:02L,destination:IDBOR, procedure path:(1st waypoint,dis to destination:162529.8,meta action:(0~45°,0~10km))...

Meta Action:1st waypoint:(0~45°,0~10km)'

******System prompt:******

'n**RNP SID Flight Procedure Path Lower Waypoint Designer** nRole:You are the SID RNP Flight Procedure Path Lower Designer, plan accurate position of next waypoint based on the meta action given by Top agent and other information. Avoid obstacles and close to destination.'n nContext:n- Coordinates: X-axis is the east, Y-axis is the north, When planning the next waypoint you are at the last waypoint as the origin(0,0).'n-Objective:plan next waypoint accurate position based on offered information.'n nInputs:1.Meta Action(IMPORTANT):the rough azimuth and distance of next waypoint, which provide reference for next accurate palming of next waypoint.'n2.destination: the azimuth and distance of the destination, you should close to it.'n3.Notable Obstacles:  ....

******User Input: ******

Meta Action:1st waypoint:(0~45°,0~10km) ndestination:(name:BOKIR,relative position:16.15°,distance:139374.8m) nnotable obstacles:(name:LL,relative position:270.71°,distance:48965.5m) nmission goal:plan 1st waypoint accurate position based on aboved information'nThink step by step.

******Assistant Output: ******

Thoughts:
-goal:plan 1st waypoint accurate position.
-meta action:1st waypoint:(0~45°,0~10km)
-notable obstacles:(name:LL,relative position:270.71°,distance:48965.5m) navoid these obstacles
-destination waypoint:(name:BOKIR,relative position:16.15°,distance:139374.8m)affect:the last few waypoints should close to the destination.
Accurate waypoint position:1st waypoint:(21.9°,3586.4m)

Fig. 2 Fine-tuning data of Plan Agent(left) and Waypoint Agent(right).

by the type of the flight procedure, and the value of which is kept constant in the main area, and gradually decreases in the sub-area.

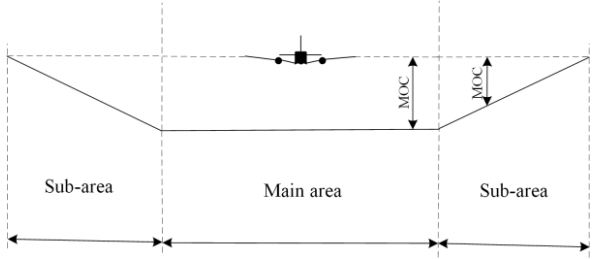

Fig. 3 The protected area of flight procedure.

*Special Constraints Compliance(SCC):* According to ICAO Doc 8168, the special limitation for a straight-line departure procedure is that the line from the runway to first waypoint does not exceed an angle of 15° from the runway direction.

*Mission Completion Rate(MCR):* It is defined that in a flight procedure design task, AutoFPDesigner makes up to 8 decisions, i.e., the agents in AutoFPDesigner have up to 8 rounds of dialogue, and the task is considered complete when the last waypoint decided upon is the termination point of the procedure.

### 4.2. Few-shot Learning

To evaluate the performance of our method, we conducted several few-shot learning experiments. We used 13.5%, 25% and 41.7% of flight procedures from Shuangliu airport in the dataset to fine-tune the Plan Agent and Waypoint Agent in AutoFPDesigner, and then tested it with the flight procedures from Shuangliu airport in the dataset that were not trained and served as Experience Memory.

The results are shown in Table 1, as the amount of fine-tuned data increases AutoFPDesigner's performance on FPS, SCC and MCR, when the percentage of fine-tuned data reaches 41.7%, AutoFPDesigner can ensure that its designed procedure fully meets the safety and specific restrictions.

Table 1 Few-shot learning

| Airport | ZUUU | | |
|---|---|---|---|
| Percentage of fine-tuning data(%) | 13.5 | 25.0 | 41.7 |
| FPS(%) | 98.18 | 100 | **100** |
| SCC(%) | 75 | 87.5 | **100** |
| MCR(%) | 37.5 | 37.5 | **75** |

In order to test the generalization performance of AutoFPDesigner, we use the fine-tuned model with the best performance in ZUUU airport directly for the test of departure procedures in the dataset of Jiangbei Airport, and the percentage of the tested procedures is 75%, and the remaining 25% of the data is used as Experience Memory in Plan Agent, and the performance performance is shown in Table 2, which can be seen from the table that when the basic airport data completely changed, AutoFPDesigner still shows good performance, which should provide a strong support for AutoFPDesigner in the actual project.

Table 2 AutoFPDesigner's performance in ZUCK

| Airport | FPS(%) | SCC(%) | MCR(%) |
|---|---|---|---|
| ZUCK | 95.83 | 85.7 | 71.4 |

### 4.3. Compatibility

We use gpt-3.5-turbo-0125 and gpt-4o as the base models of each Agent in AutoFPDesigner, and test them procedure design of ZUUU in the dataset, and their performance is shown in Table 3 , from which we can see that when the base model is changed, AutoFPDesigner still shows a good performance, which proves the compatibility of AutoFPDesigner for different base LLM models.

Table 3 Compatibility to different LLMs

| Base LLM | FPS(%) | SCC(%) | MCR(%) |
|---|---|---|---|
| GPT-3.5-Turbo-0125 | 100 | 100 | 75 |
| GPT-4o | 100 | 100 | 66.7 |

### 4.4. Stability

LLM is usually disturbed by hallucination, which are detrimental to AutoFPDesigner, in order to study this effect we have performed stability tests on AutoFPDesigner and we have performed the following expeiments：

**Experiment 1:** Frequency of no hallucination for Plan Agent and Waypoint Agent with varying amounts of fine-tuning data, two key agents in the design of flight procedures, whose stability is critical to AutoFPDesigner's performance, with a test set of unused procedures at Shuangliu Airport in the dataset.

**Experiment 2:** The stability of all agents work in AutoFPDesigner, the test set is the unused procedures in the Shuangliu and Jiangbei airports part of the dataset, they are about 30% of the total dataset.

The statistical results of Experiment 1 are shown in the Fig. 4 Plan Agent and Waypoint Agent are tested about 50 times in different task scenarios, and their performances have good stability, especially when the percentage of fine-tuning data is 41.7%, Plan Agent and Waypoint Agent achieve zero invalid output.

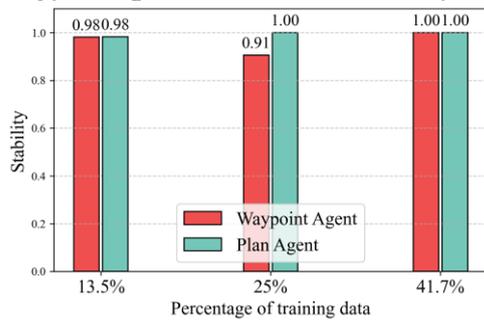

Fig. 4 Stability of Waypoint Agent and Plan Agent.

The statistical results of Experiment 2 are shown in the Fig. 5 after about 90 tests for all agents in different task scenarios, except for Task Manager they all showed different degrees of hallucination, as well as Plan Agent and Waypoint, which were fine-tuned with 41.7% of the data. We believe that this is the result of not considering the dialogue of other agents in the fine-tuning process.

We analysed Agents chat transcripts in Experiment 2, and we found that the outputs of most fantasy scenarios were agent-acceptable, agents understood and judged the statements made by other agents, e.g., the Waypoint Agent's most likely fantasy scenario is shown Fig. 6. where the Waypoint Agent loses planning Waypoint serial number information, but the Calculate Agent usually gets the latitude and longitude of the previous Waypoint correctly based on the dialogue transcript. We believe that this ability of self-judgement between multi agents is the key to achieving high performance in AutoFPDesigner.

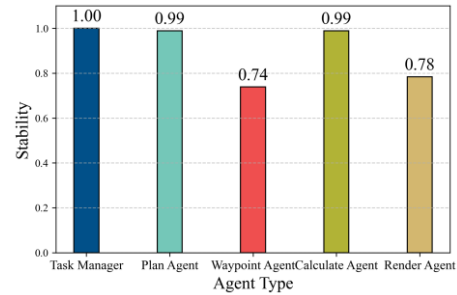

Fig. 5 Stability of agents in AutoFPDesigner.

**Finished waypoints:**[[30.672785,103.991117],[30.709621,104.008689],[30.820674,104.13808]]

**Plan Agent**

Thoughts:
-is finished planning:the distance to destination is 124358.4,not arrive destination!
-planning goal:plan 4th waypoint relative position from sid runway:02L to GURET.
notable obstacles:No notable obstacles navoid these obstacles
similar sid procedure path from same sid runway: sid runway:02L,destination:IDBOR,procedure path:(1st waypoint,dis to destination:162529.8,meta action:(0−45°,0-10km)),(2nd waypoint,dis to destination:164619.6,meta action:(0−45°,0-10km)),(3rd waypoint,dis to destination:170508.1,meta action:(45−90°,10-20km)),(4th waypoint,dis to destination:172247.2,meta action:(135−180°,20-30km)),(5th waypoint,dis to destination:144245.9,meta action:(180−225°,10-20km)),(6th waypoint,dis to destination:132674.0,meta action:(135−180°,50+km)),(7th waypoint:arrival destination!) affect the planning of the initial few waypoints can be provided for reference,if none,try to fit first waypoint plan restriction and avoid notable obstacles.
similar sid procedure path from same destination: sid runway:20L,destination:GURET,procedure path:(1st waypoint,dis to destination:153776.5,meta action:(180−225°,0-10km)),(2nd waypoint,dis to destination:156593.7,meta action:(180−225°,20-30km)),(3rd waypoint,dis to destination:172027.2,meta action:(90−135°,10-20km)),(4th waypoint,dis to destination:164116.8,meta action:(0−45°,20-30km)),(5th waypoint,dis to destination:140993.8,meta action:(0−45°,30-50km)),(6th waypoint,dis to destination:100634.1,meta action:(45−90°,50+km)),(7th waypoint:arrival destination!)
affect the planning of the last few waypoints can be provided for reference, if none,last few waypoints try to close destination.
-destination waypoint:(name:GURET,relative position:68.04°,distance:124358.4m)affect the last few waypoints should close to the destination.
**Meta Action:** 4th waypoint:(0−45°,20-30km)

**Waypoint Agent**

Thoughts:
-goal:plan (0− waypoint accurate position.
-meta action:(0−45°,20-30km)
notable obstacles: navoid these obstacles
-destination waypoint:name:GURET,relative position:68.04°,distance:124358.4m,affect the last few waypoints should close to the destination.
**Accurate waypoint position:**(27.4°,25700.7m)

**Calculate Agent**

Thoughts:
"arguments": "{"origin_lat":30.820674, "origin_lon":104.13808, "bearing":27.4, "distance":25700.7}", "name":
(Use Calculator tool)
[31.025817, 104.262205]

Fig. 6 Acceptable hallucination communication.

Table 4 Ablation of components in AutoFPDesigner

| ID | Experience Memory | Chain of Thought | Multi-Agent | FPS(%) | SCC(%) | MCR(%) |
|---|---|---|---|---|---|---|
| 1 | ✗ | ✓ | ✓ | 100 | 100 | 25 |
| 2 | ✓ | ✗ | ✓ | 100 | 0 | 12.5 |
| 3 | ✓ | ✓ | ✗ | 95.83 | 62.5 | **100** |
| 4 | ✓ | ✓ | ✓ | **100** | **100** | 75 |

### 4.5. Interpretability

Unlike flight procedure design methods based on heuristic algorithms or traditional neural network algorithms, AutoFPDesigner possesses the good interpretability property of LLM. As shown in Fig. 6 AutoFPDesigner's decision-making processes are all output through natural language, and the whole process is transparent and interpretable.

### 4.6. Ablation Experiment

Table 4 shows the results of ablating different components in AutoFPDesigner. The base llm used for all agents is gpt-3.5-turbo-0125, and the Plan Agent and Waypoint Agent use the model fine-tuned with 41.7% data. From ID1 to ID3, we disabled Exp Memory, CoT and Multi Agent in AutoFPDesigner. In the case of Multi Agent, we combined Plan Agent and Waypoint Agent into one agent, which directly outputs the precise bearing and distance of the next waypoint.

Exp Memory removal reduced AutoFPDesigner's task completion rate, and CoT removal caused a severe decrease in task completion rate and in meeting the special limits of the procedure. Combining the Plan Agent and Waypoint Agent into one Agent decreases the FPS and SCC performance and enhances the MCR performance of AutoFPDesigner.

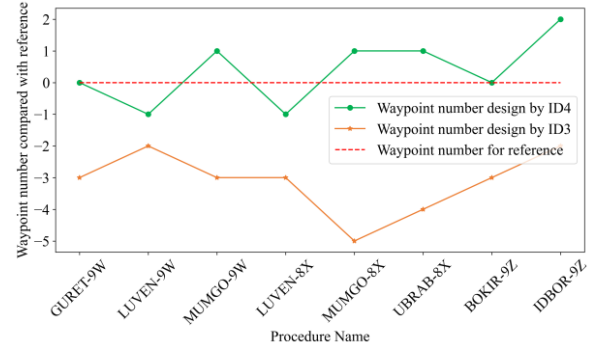

Fig. 7 Comparison of the number of waypoints for experiments ID3 and ID4.

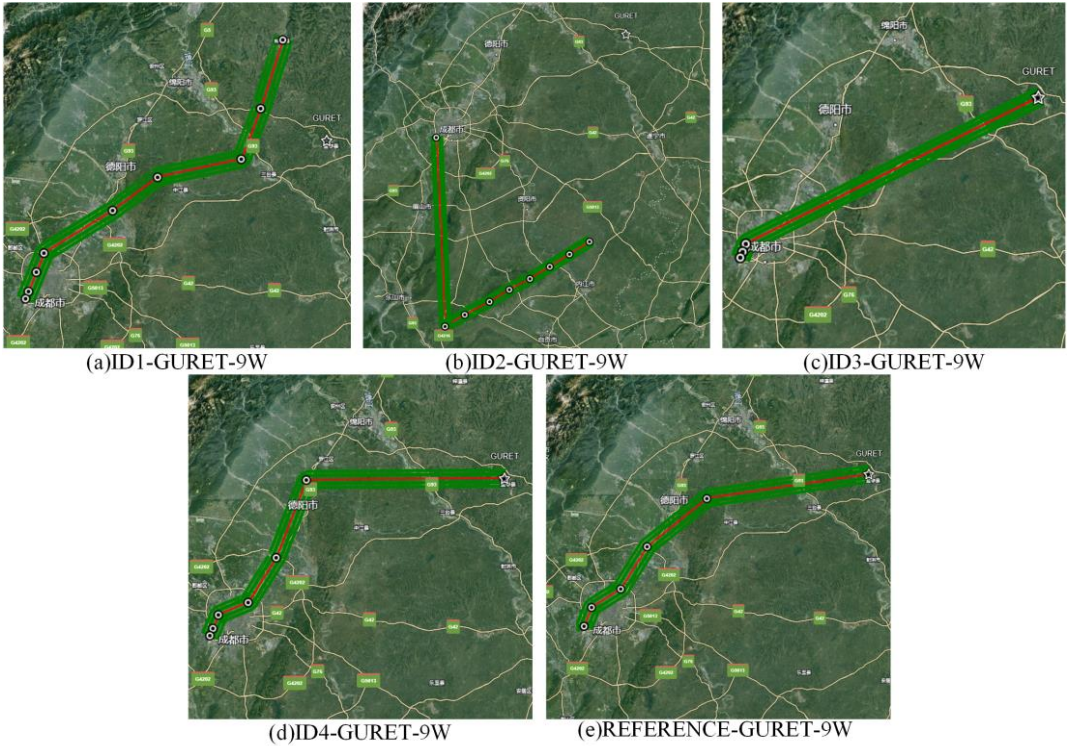

(a)ID1-GURET-9W  (b)ID2-GURET-9W  (c)ID3-GURET-9W

(d)ID4-GURET-9W  (e)REFERENCE-GURET-9W

Fig. 8 Procedure GURET-9W render in ID1-4 and reference.

We compare the waypoint number in the flight procedures designed by ID3 and ID4 with the actual number of waypoints in the corresponding procedures in the dataset, as shown in Fig. 7 which shows that AutoFPDesigner without multi agent prefers to complete the design with fewer waypoints, i.e., by connecting the start point and the end point directly, which has the advantage of higher MCR. The advantage of doing so is that the MCR is higher, but the disadvantage is that the safety and the index of meeting the special constraints are reduced.

We visualized the GURET-9W procedure from ID1-4 and compared them with the reference procedures in the dataset. As shown in Fig. 8, compared with GU-RET-9W in ID1-3, GURET-9W in ID4 is better similarity to the reference procedure. We present the visual results of all flight procedure design experiments in Appendix A3. It can be observed that the situation depicted in Fig. 8 is not an isolated case. This suggests that some of the design styles of AutoFPDesigner are similar to those of professional designers.

*4.7. Human involved feedback*

We conducted human involved experiments with AutoFPDesigner, where users provided feedback on the design results in the form of natural language. This feedback was based on the reasoning process and visualization results during the AutoFPDesigner design process. The purpose of this experiment is to demonstrate that AutoFPDesigner has the capability to collaboratively design flight procedures with human users, This is also an indispensable part of the LLM agent-led flight procedure design framework.

In this section experiments we co-design with AutoFPDesigner the flight procedures contained in the Shuangliu Airport test set in the dataset, and we set the human feedback to be of two types：

1. no fix: No need to adjust the procedure designed by AutoFPDesigner.

2. fix: [fix_waypoint, bearing, distance, last_waypoint_lat, last_waypoint_lon].

fix_waypoint: which waypoint need to fix, this parameter is required.

bearing: the fix waypoint relative bearing to last waypoint, this parameter is required.

distance: the fix waypoint distance to last waypoint, his parameter is required.

last_waypoint_lat: the last waypoint latitude of fix waypoint, this parameter is optional.

last_waypoint_lon: the last waypoint longitude of fix waypoint, this parameter is optional.

We recommend adding more information such as the latitude and longitude of the last waypoint to reduce the hallucination that AutoFPDesigner may experience when understanding user feedback. Based on the above experimental settings with human feedback, Table 5 show that the flight procedure designed with human user participation can achieve better results compared to the flight procedure designed by AutoFPDesigner solely.

Table 5 Performance of AutoFPDesigner with human assistance

|  | FPS(%) | SCC(%) | MCR(%) |
|---|---|---|---|
| No human feedback | 100 | 100 | 75 |
| human feedback | **100** | **100** | **100** |

## 5.Discussion and future work

In our work, despite AutoFPDesigner demonstrating excellent performance in safety, meeting special constraints, and efficiency, we believe there are several points worth discussing and exploring:

(1) Comparison with Heuristic and Machine Learning-Based Flight Procedure Design Methods and Commercial Software: This paper does not include a quantitative comparison with heuristic and machine learning-based flight procedure design methods or commercial flight procedure design software. However, we believe that AutoFPDesigner's user-friendliness is unmatched by other methods. Heuristic and machine learning-based methods suffer from significant generalization issues and require specialized algorithm developers for modeling, which we consider a major limitation to their practical application. In future work, we plan to compare AutoFPDesigner's performance with that of commercial flight procedure design software and explore the possibility of integrating AutoFPDesigner as a plugin into existing commercial flight procedure design tools.

(2) API Costs: The experiments in this study utilized OpenAI's API, costing approximately \$120. The token cost for designing a single flight procedure with AutoFPDesigner is approximately \$0.003. These costs provide a reference for future researchers.

(3) Incorporation of RAG into AutoFPDesigner: AutoFPDesigner understands flight procedure design specifications through Rules; however, the design specifications within Rules still require manual filtering. We believe that integrating RAG (Retrieval-Augmented Generation) with appropriate prompts could enable AutoFPDesigner to extract relevant information from flight procedure specifications directly based on design tasks.

## 6.Conclusion

The work introduces AutoFPDesigner, an innovative approach powered by large language models (LLMs) for flight procedure design, fundamentally transforming the traditional expert-driven design process. Our research has achieved significant advancements in this field, which are summarized as follows:

(1) Through the integration of Rules & Experience Memory, Tool Library, and multi-agent architecture, AutoFPDesigner is equipped with professional capabilities in the field of flight procedure design.

AutoFPDesigner can comprehend user requirements via natural language interaction and autonomously complete the design process, while allowing seamless user intervention throughout the design workflow.

(2) We conducted automated design experiments for departure procedures at Shuangliu and Jiangbei airports using AutoFPDesigner, followed by a comprehensive analysis of its generalization ability, stability, and interpretability. The results indicate that AutoFPDesigner demonstrates strong generalization, stability, and interpretability, with the designed departure procedures nearly fully complying with safety requirements.

(3) We consider our work a crucial step toward automated flight procedure design. For future work, we aim to enhance AutoFPDesigner by integrating Retrieval-Augmented Generation (RAG) to enable the direct reading of flight procedure specifications. Additionally, we are exploring the potential integration with commercial flight procedure design software to further expand AutoFPDesigner's professional capabilities.

# Appendix

## A1 The prompt of agents in AutoFPDesigner

We utilize AutoGen[22] to implement the multi-agent architecture. According to AutoGen, the use of tools is managed through two agents: one agent extracts input parameters for the tool, while the other agent execute the tool and returns the results. In AutoFPDesigner, the fine-tuned Plan Agent llm and Waypoint Agent llm are considered "tools" with attributes of large language models. The prompts for these agents are as follows:

**Group manager:** Group chat manager.

**Task agent:** You are the task manager for SID flight procedure design, and you have two tasks: task 1:extract useful information from user's input, for example, the runway name(such as RWY02R), destination. task2:just tell other agents design next waypoint until WaypointDesign_Top_Exe output finished design!do not give or summary other information!do not give juegement wheather finished design!

**Plan agent(input):** you are the sid procedure top design agent input transfer. The input of WaypointTopAgentModel function is as follows: sid_runway(str):sid_runway name,destination(str):destination name so_far_waypoints(list[float]):the design finished waypoint longitude and latitude [[lat1 (degree), lon1 (degree)], [lat2 (degree), lon2 (degree)],...], if no so far waypoints, so_far_waypoints: [];Don't give judgement wheather finished design! just summary the input! Only call WaypointTopAgentModel once per conversation. waypoint_start_lat(float): if so_far_waypoints is empty,it is runway latitude, else,the final waypoint latitude of so_far_waypoints,waypoint_start_lon(float):if so_far_waypoints is empty,it is runway longtitude, else,the final way waypoint longtitude of so_far_waypoints, when design 1st waypoint , so_far_waypoints is empty, waypoint_start_lat is runway latitude, waypoint_start_lon is runway longtitude,when design 2nd waypoint, so_far_waypoints include 1st waypoint latitude and longtitude, waypoint_start_lat is 1st waypoint latitude, waypoint_start_lon is 1st waypoint longtitude and so on.

**Plan agent(exe):** \n**RNP SID Flight Procedure Path Top Designer**\nRole:You are the SID RNP Flight Procedure Path Top Designer, plan relative position and distance of next waypoint in sequence based on offered information. Avoid obstacles and close to destination. \n\nContext\n- Coordinates: X-axis is the east, Y-axis is the north, When planning the next waypoint you are at the last waypoint as the origin(0,0), heading=0 is north.\n\n-first waypoint restriction: The direction of the first waypoint should not exceed an angle of 15 degrees from the runway heading.\n -similar procedure: the similar sid runway procedure and the similar destination procedure can offer advice for your plan.\n-Objective:plan next waypoint in sequence based on offered information.\n\nInputs\n1.the planning sid procedure name:Unique identifier of sid procedure.\n2. sid runway: the name of sid runway and runway heading.\n3. destination: the name of terminal waypoint and relative position information.\n4. similar procedure: the most similar procedure path from same sid runway :Affect the planning of the initial few waypoints.\n5.the most similar procedure path from same destination:Affect the planning of the last few waypoints\n6.Notable Obstacles: the notable obstacles in 50 kilometers, which you should avoid them.\n historical planning:finshed waypoints\n7.Mission Goal:plan next waypoint based on offered information.\n Output:\n1.Thoughts:Write down the process of your thinking.\n2.Meta Action(MOST IMPORTANT): Relative position and distance of the next waypoint or finish sid procedure path design.

**Waypoint agent(input):** you are the sid procedure waypoint design WaypointLowAgentModel input transfer. the input of WaypointLowAgentModel function is as follows: meta_action(str):the roughly plan of next waypoint,for example:(1st waypoint:(northeast,0-10km));destination(str):destination name and relation and postition,for example:(name:LUVEN,relative position:176.73°,distance:134660.8m);obstacles(str): the name and postion of obstacles, if arrival destination,you just tell other agents finished the design. do not give or summary other information. Call WaypointLowAgentModel only once per conversation.

**Waypoint agent(exe):** \n**RNP SID Flight Procedure Path Lower Waypoint Designer**\nRole:You are the SID RNP Flight Procedure Path Lower Designer, plan accurate position of next waypoint based on the meta action given by Top agent and other information. Avoid obstacles and close to destination. \n\nContext\n- Coordinates: X-axis is the east, Y-axis is the north, When planning the next waypoint you are at the last waypoint as the origin (0,0).\n-Objective:plan next waypoint accurate position based on offered information.\n\nInputs:1.Meta Action(IMPORTANT):the rough azimuth and distance of next waypoint, which provide reference for next accurate palnning of next waypoint. \n2. destination: the azimuth and distance of the destination, you should close to it. \n3. Notable Obstacles: the notable obstacles in 50 kilometers, which you should avoid them.\n4.Mission Goal:plan next waypoint accurate position based on offered information\n Outputs:1.Thoughts:Write down the process of your thinking.\n2.Accurate waypoint position:the accurate aazimuth!! and distance of next waypoint!! (Important)only only only one waypoint each time!!!

**Calculate agent:** The function inputs as follows:origin_lat(float):the origin latitude,for 1st waypoint design is the runway latitude,for 2nd waypoint design is the 1st waypoint latitude and so

on;origin_lon(float):the origin longitude,for 1st waypoint design is the runway longitude,for 2nd waypoint design is the 1st waypoint longitude and so on;bearing(float):the azimuth of the next waypoint,provided by WaypointDesignAgent,the unit of angle is degree;distance(float):the distance of the next waypoint,provided by WaypointDesignAgent,the unit of distance is m, for example:{'origin_lat': 30.593333, 'origin_lon': 103.954167, 'bearing': 21.9, 'distance': 3551.4},summary the inputs as example format!! Not calculate!!! transfer inputs to calculate_new_lat_lon function!!!!!

**Render agent:** You are the input transfer of RenderAgent function. The function inputs as follows:sid_runway_name(str): the name of sid runway,for example,RWY02R and so on;so_far_waypoints(list):the design finished waypoint longitude and latitude information.[[lat1 (degree), lon1 (degree)], [lat2 (degree), lon2 (degree)],...],not include runway init lat and lon,if no so far waypoints, so_far_waypoints: [];destinatin(str):the name of destiantion

**Fixwaypoint agent:** You are the input transfer of fixwaypointbyHF function,get the inputs from other agents. The function inputs as follows:action(str):fix or no fix,if fix, other inputs are needed;if no fix, other inputs can be null. so_far_waypoints:the design finished waypoint longitude and latitude information.[[lat1 (degree), lon1 (degree)], [lat2 (degree), lon2 (degree)]...],include need fix waypoint last_waypoint_lat(float):the last waypoint latitude of fix waypoint;last_waypoint_lon(float):the last waypoint longitude of fix waypoint;fix_bearing(float):the bearing hunman feedback(float);fix_distance(float):the distance hunman feedback(float); just get the inputs of fixwaypointbyHF function,do not give or summary other information. the input of user will like:no fix or 1stwaypoint fix, fix_bearing:21.9, fix_distance:3551.4.

## A2 Evaluation Metrics

In this section, we provide more details for calculate FPS, SCC and MCR.

FPS can be calculated by:

$$FPS = \frac{1}{N_{leg}}\sum_{i=1}^{N_{leg}} leg_{safe} \qquad (3)$$

Where $N_{leg}$ is the total number of flight procedure leg in test, $\sum_{i=1}^{N_{leg}} leg_{safe}$ represents the number of safe flight procedure leg.

SCC can be calculated by:

$$SCC = \frac{1}{N_{procedure}}\sum_{i=1}^{N_{procedure}} P_{\theta<15°} \qquad (4)$$

Where $N_{procedure}$ is the total number of flight procedure in test, $\sum_{i=1}^{N_{procedure}} P_{\theta<15°}$ represents the number of procedure these meet first waypoint does not exceed an angle of 15° from the runway direction.

MCR can be calculated by:

$$MCR = \frac{1}{N_{procedure}}\sum_{i=1}^{N_{procedure}} P_{finish} \qquad (5)$$

Where $N_{procedure}$ is the total number of flight procedure in test, $\sum_{i=1}^{N_{procedure}} P_{finish}$ represents the number of procedure these last waypoint is the termination point of the procedure within 8 communication rounds.

In this section, we provide a visual comparison between the flight procedures designed by AutoFPDesigner and the corresponding reference procedures in the dataset.

## A3 Procedure Render

### A3.1 reference procedures in the dataset

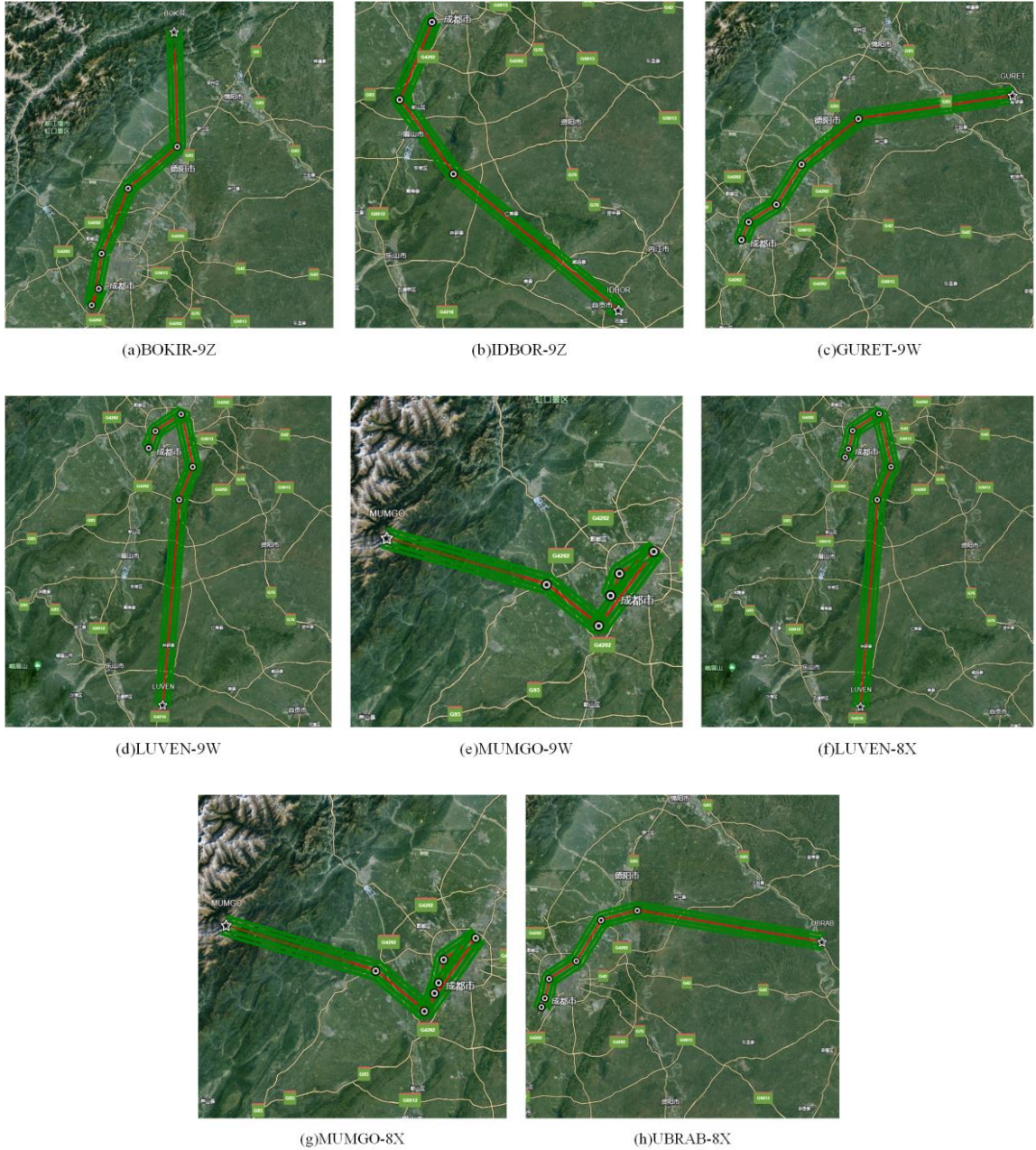

(a)BOKIR-9Z    (b)IDBOR-9Z    (c)GURET-9W

(d)LUVEN-9W    (e)MUMGO-9W    (f)LUVEN-8X

(g)MUMGO-8X    (h)UBRAB-8X

Fig. 9 reference procedures in ZUUU

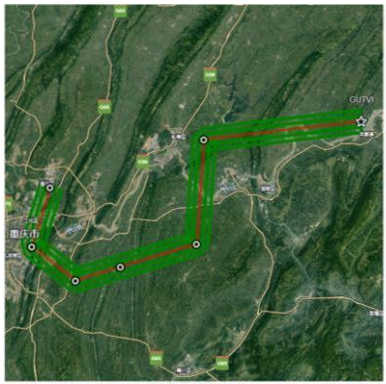
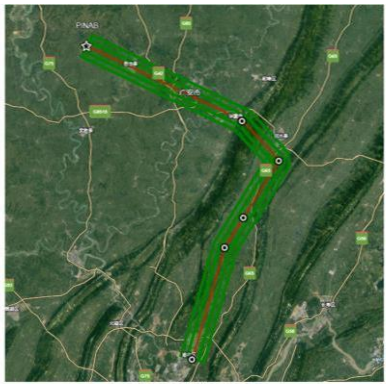
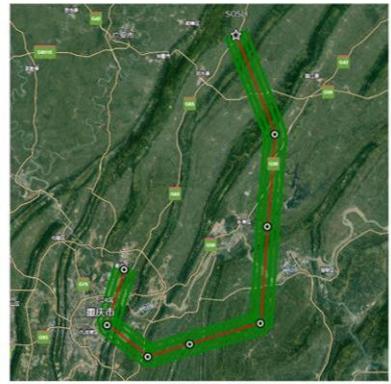

(a)GUTVI-2Y       (b)PINAB-2Y       (c)SOSLI-2Y

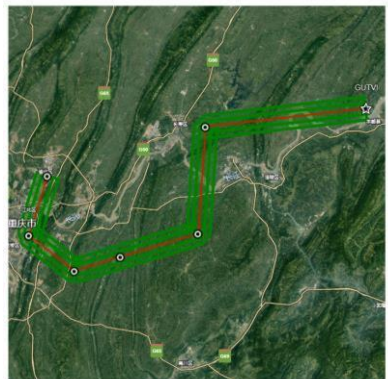
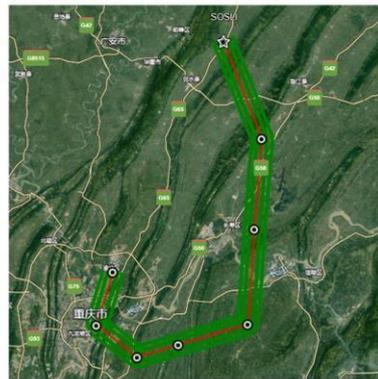
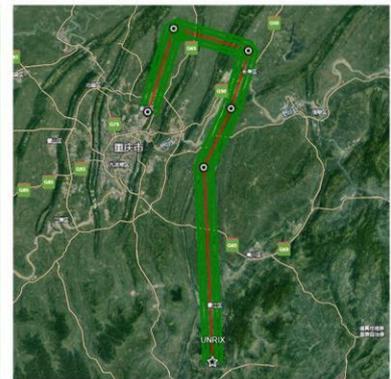

(d)GUTVI-1Y       (e)SOSLI-1Y       (f)UNRIX-1Y

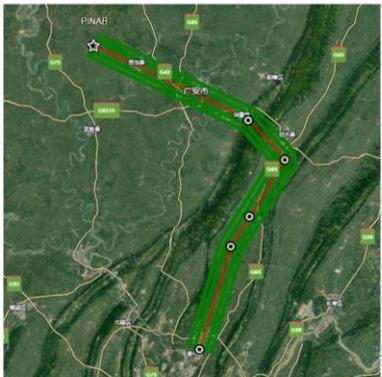
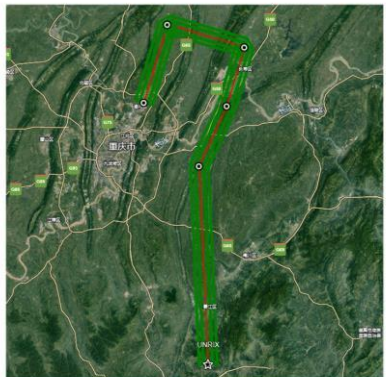
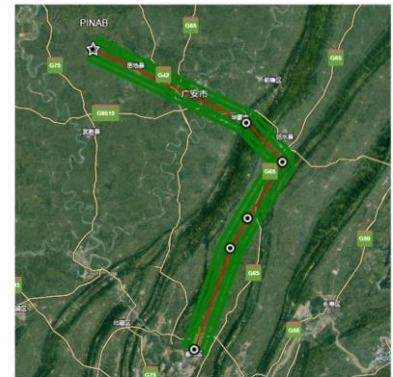

(g)PINAB-1Z       (h)UNRIX-1Z       (i)PINAB-2Z

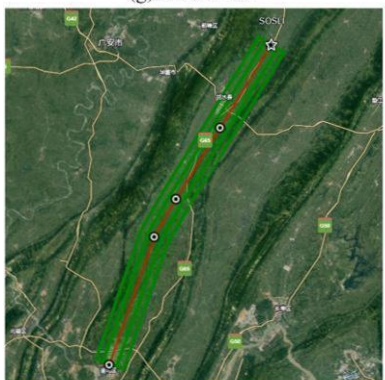
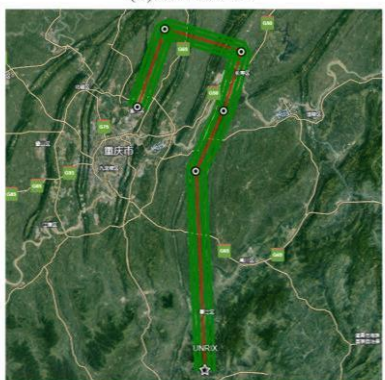
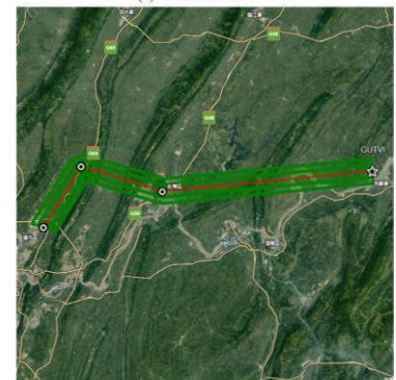

(j)SOSLI-2Z       (k)UNRIX-2Z       (l)GUTVI-3Z

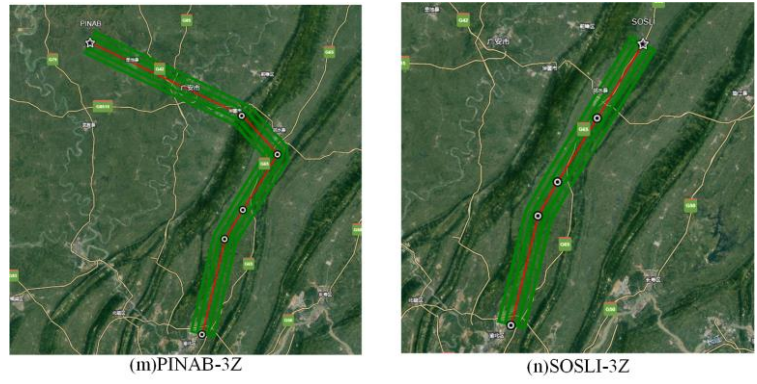

(m)PINAB-3Z                  (n)SOSLI-3Z

Fig. 10 reference procedures in ZUCK

**A3.2 procedure render in few-show learning experiment**

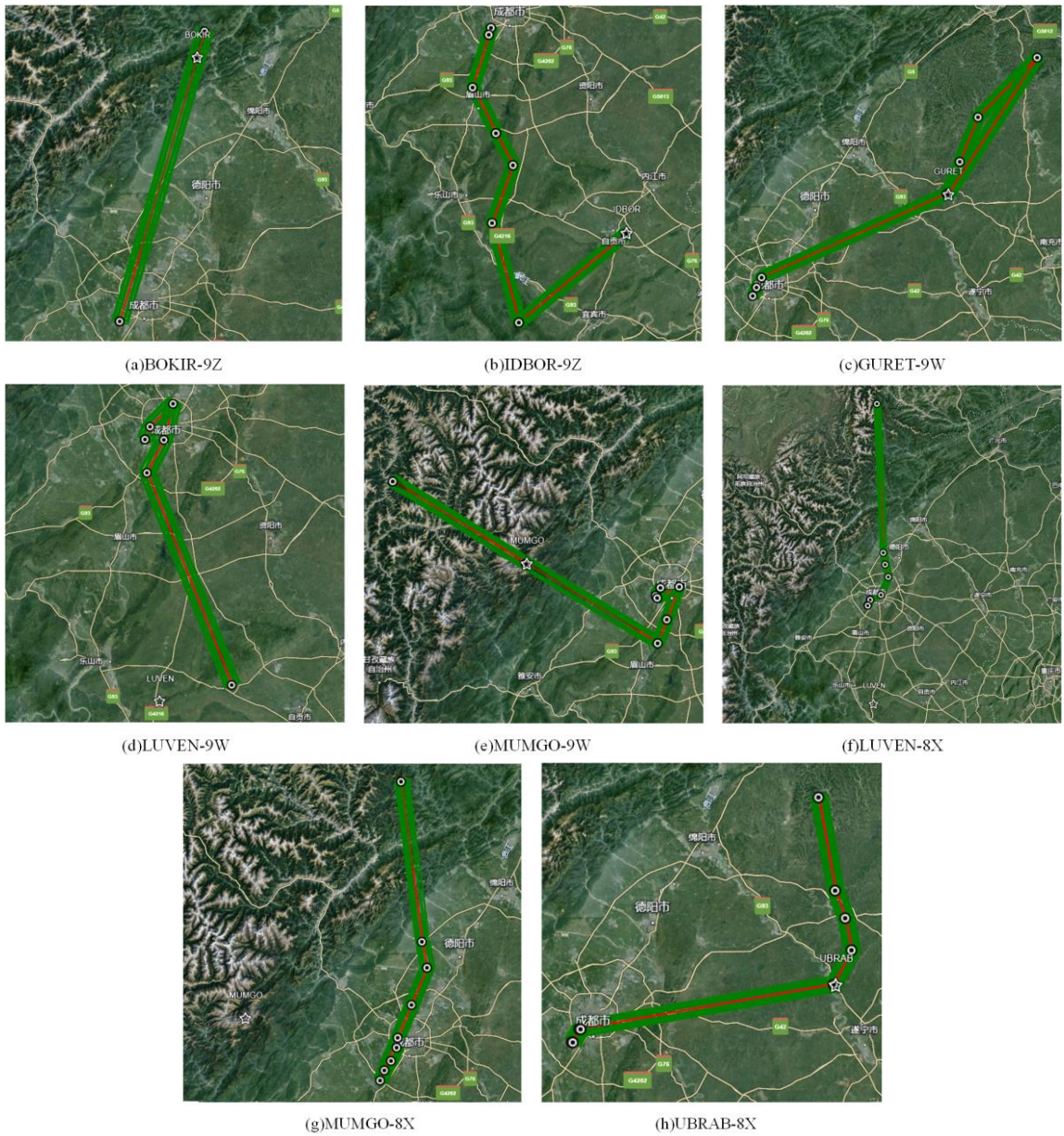

(a)BOKIR-9Z                  (b)IDBOR-9Z                  (c)GURET-9W

(d)LUVEN-9W                  (e)MUMGO-9W                  (f)LUVEN-8X

(g)MUMGO-8X                  (h)UBRAB-8X

Fig. 11 Procedures render of 13.5% fine-tuning data

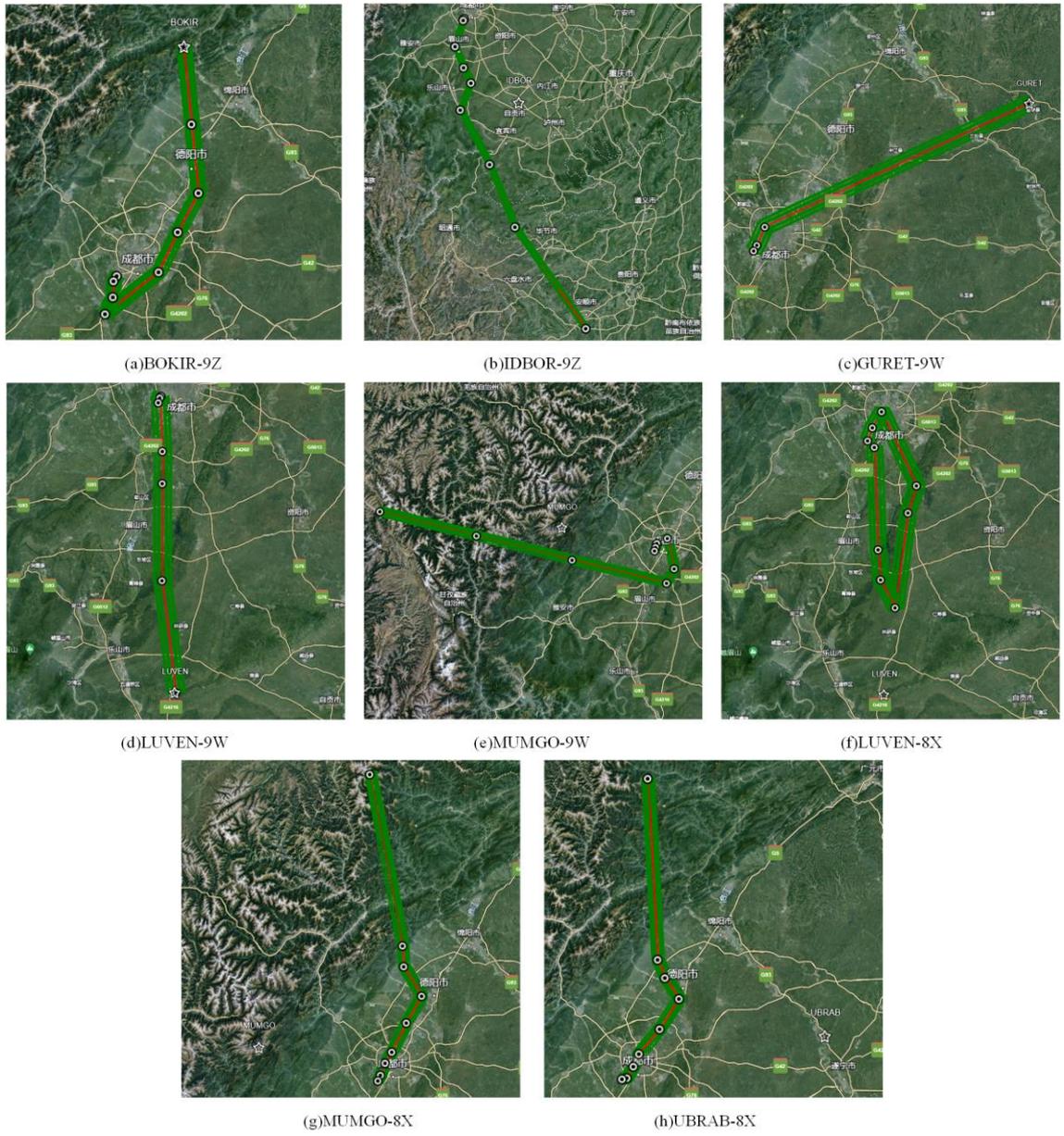

Fig. 12 Procedures render of 25.0% fine-tuning data

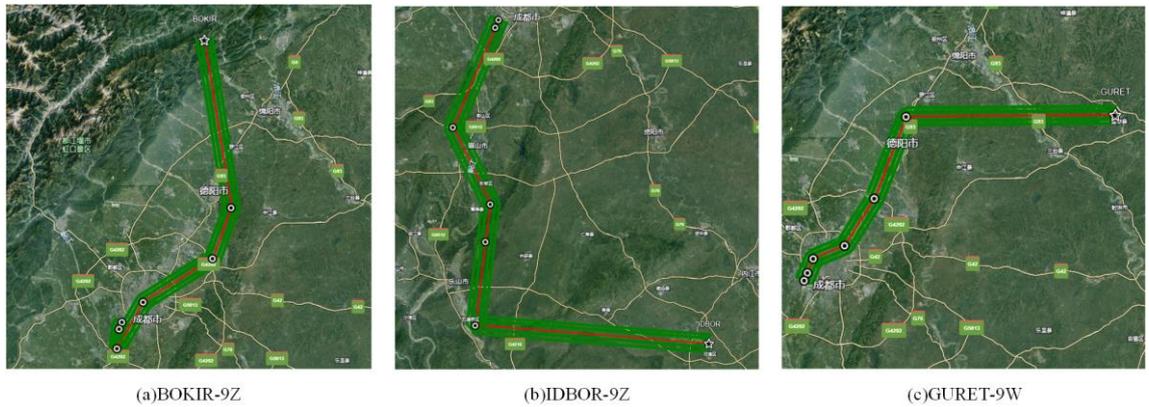

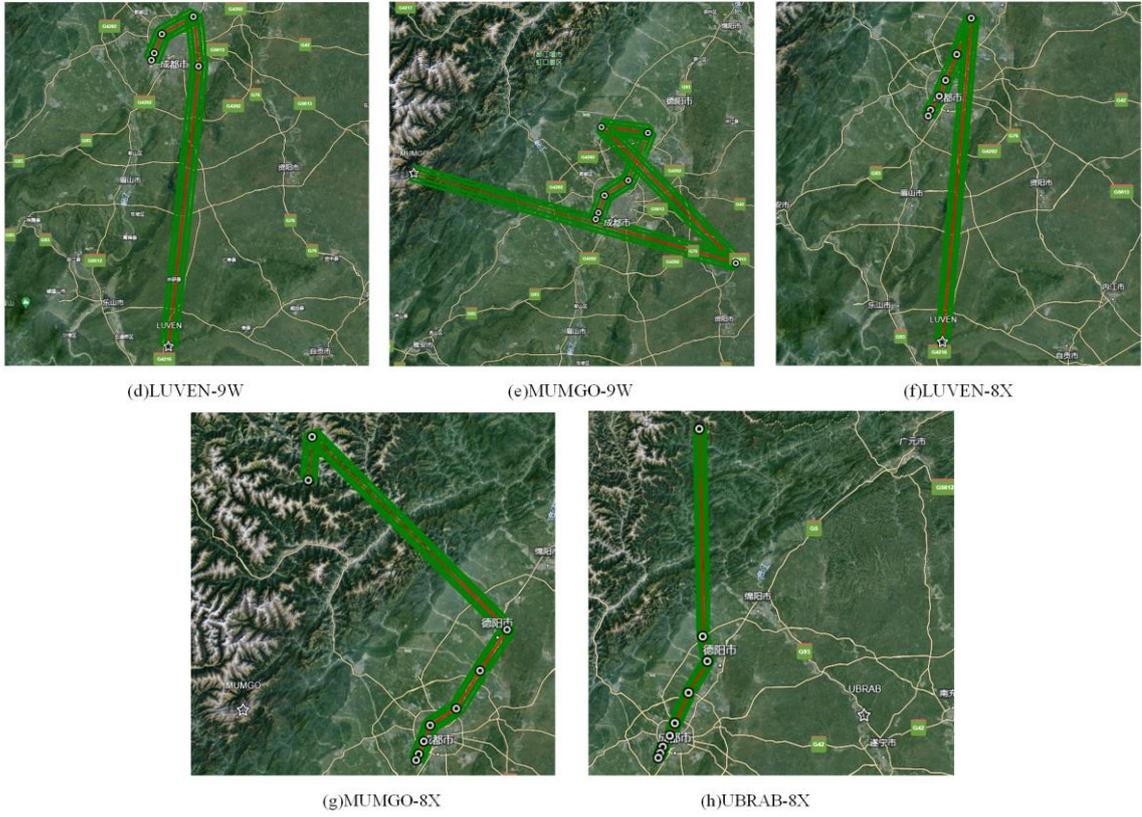

(d)LUVEN-9W           (e)MUMGO-9W           (f)LUVEN-8X

(g)MUMGO-8X           (h)UBRAB-8X

Fig. 13 Procedures(ZUUU) render of 41.7% fine-tuning data

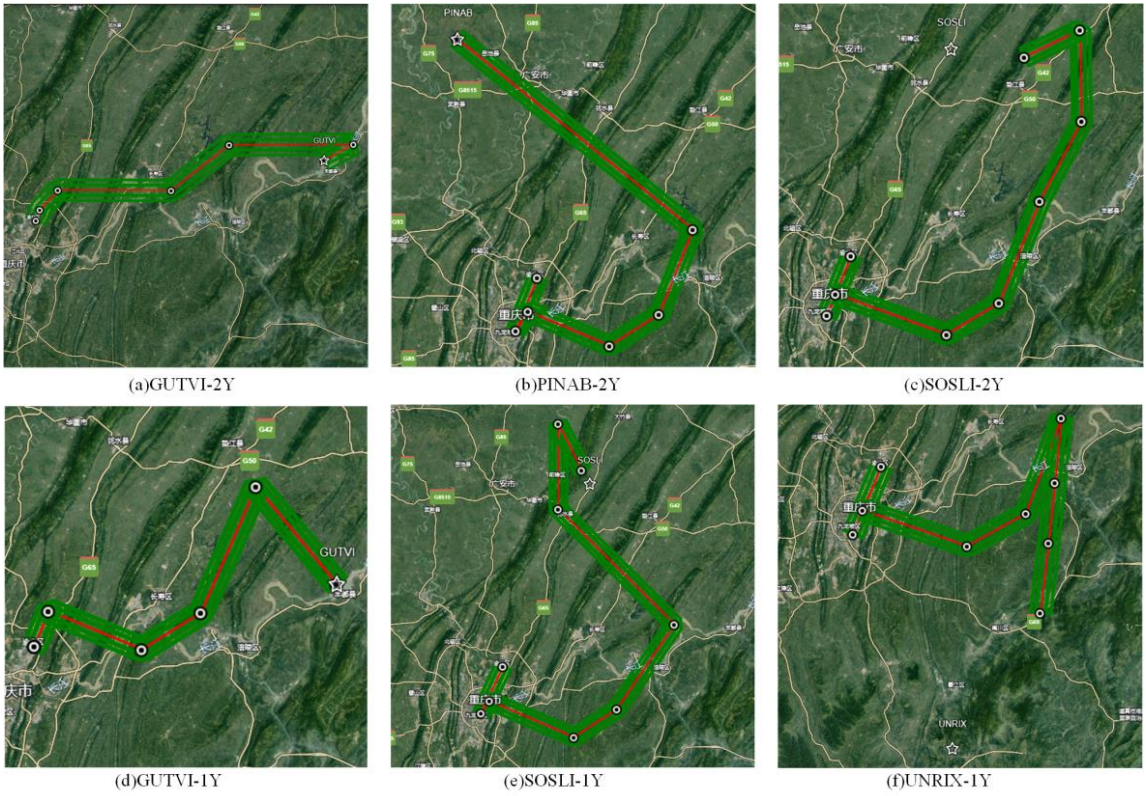

(a)GUTVI-2Y           (b)PINAB-2Y           (c)SOSLI-2Y

(d)GUTVI-1Y           (e)SOSLI-1Y           (f)UNRIX-1Y

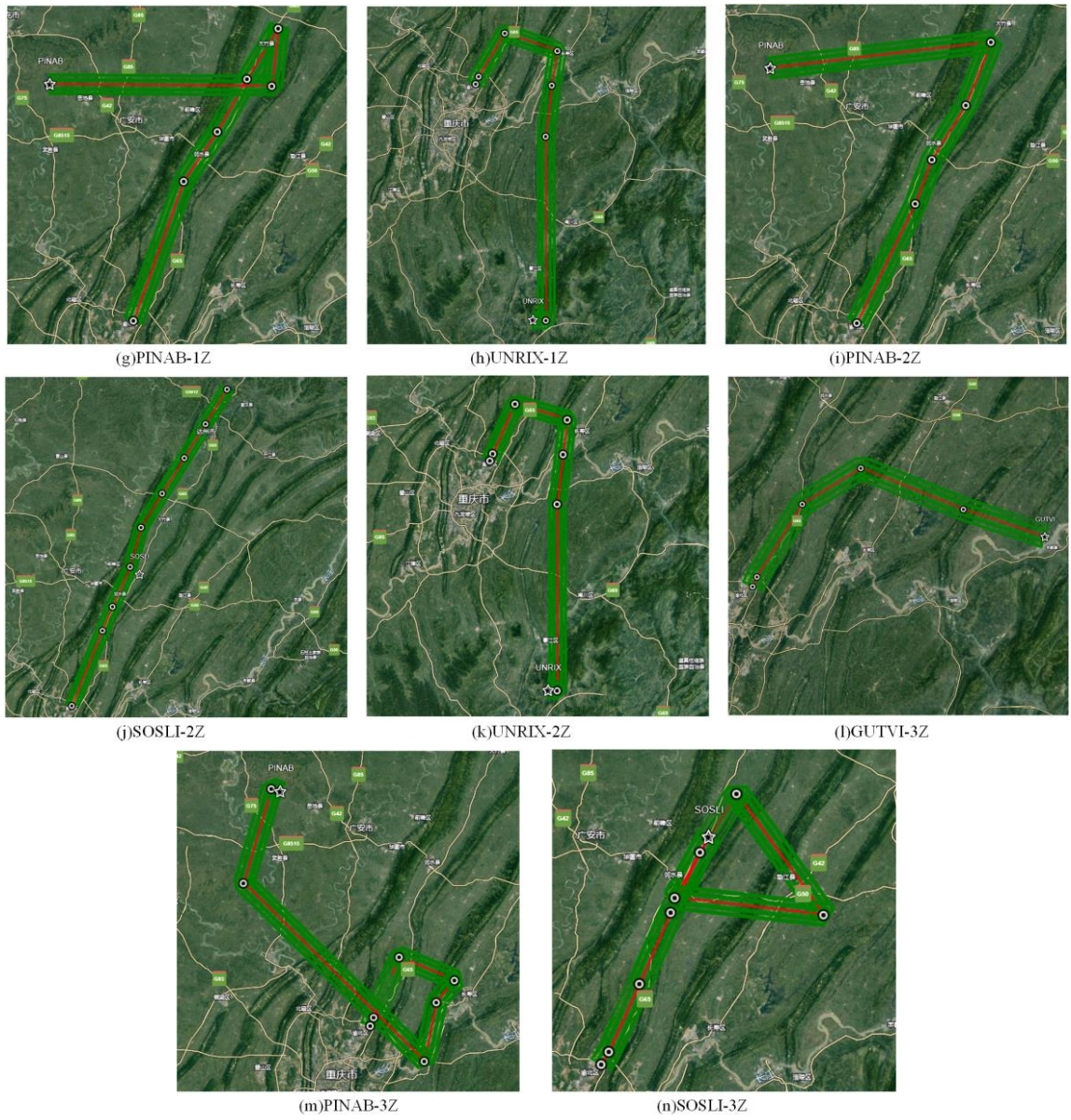

Figure 14 Procedures(ZUCK) render of 41.7% fine-tuning data

## A3.3 procedure render in compatibility experiment

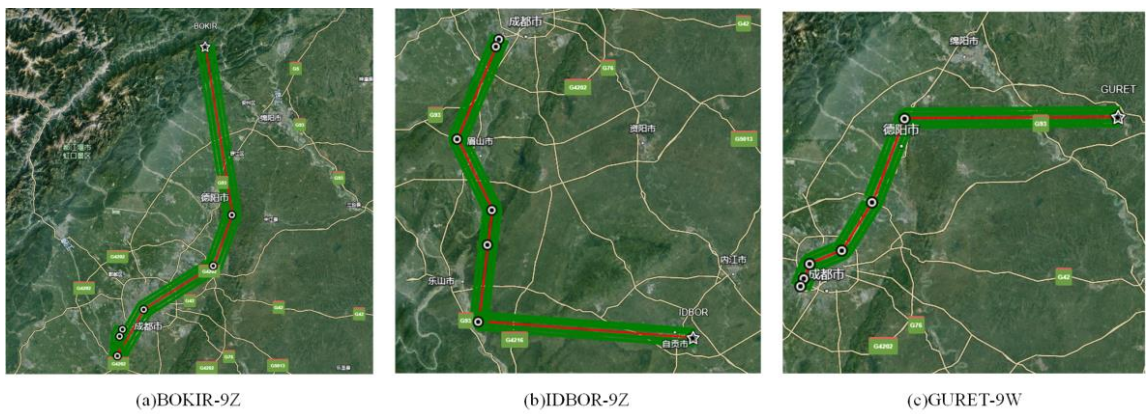

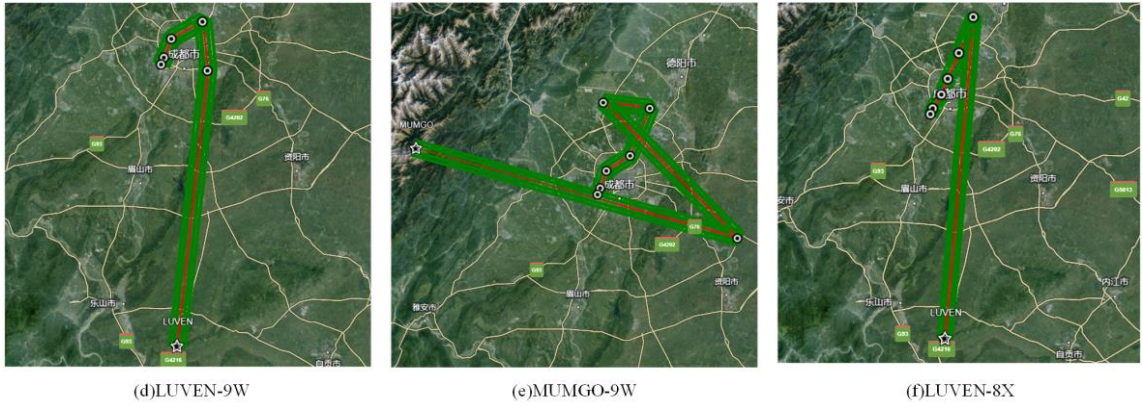

(d)LUVEN-9W · · · · · · · · · · (e)MUMGO-9W · · · · · · · · · · (f)LUVEN-8X

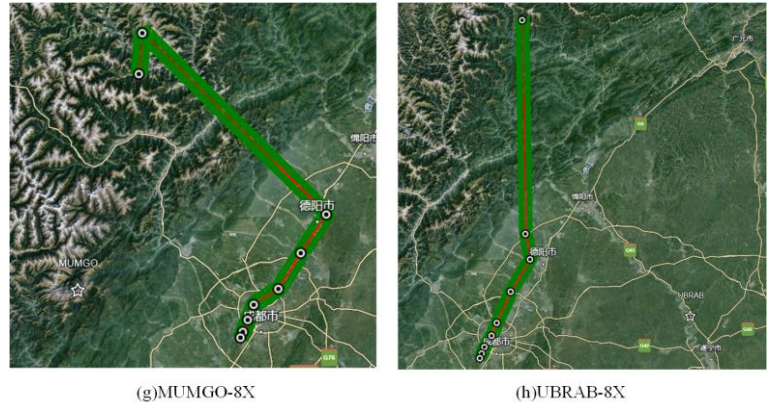

(g)MUMGO-8X · · · · · · · · · · (h)UBRAB-8X

Fig. 15 Procedures render based on gpt-3.5

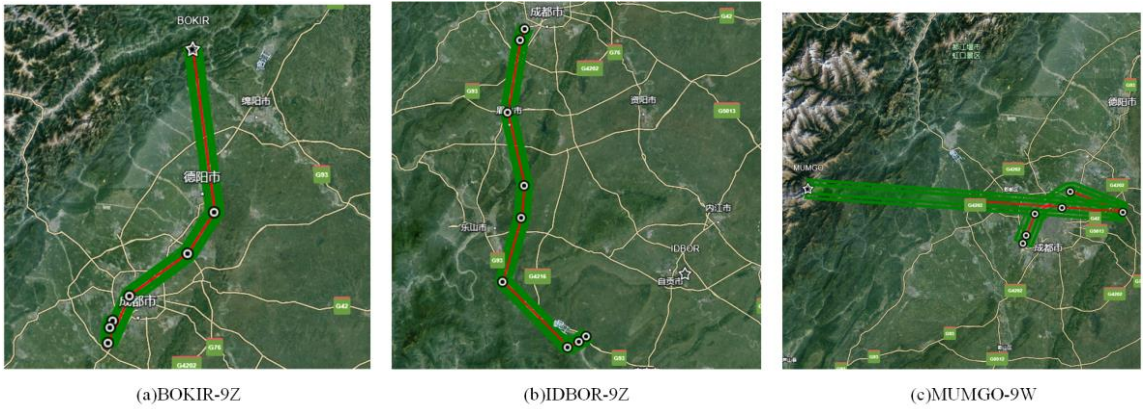

(a)BOKIR-9Z · · · · · · · · · · (b)IDBOR-9Z · · · · · · · · · · (c)MUMGO-9W

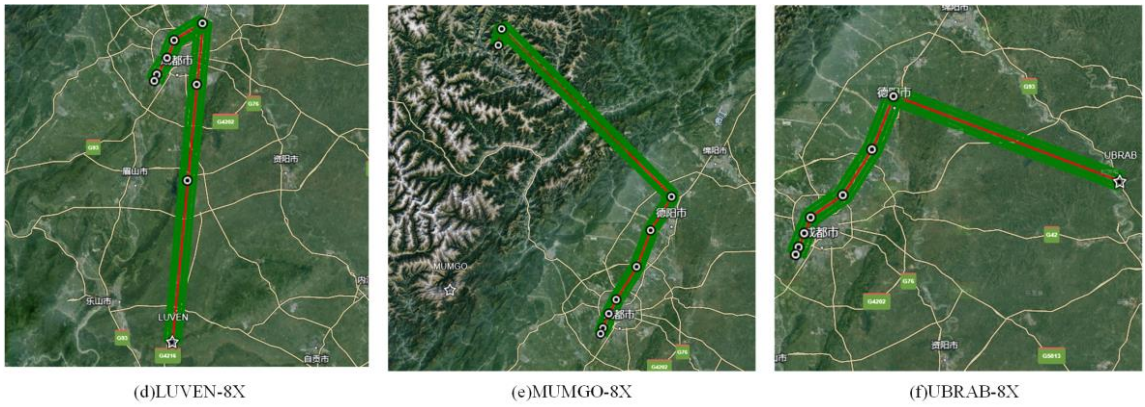

<div align="center">(d)LUVEN-8X      (e)MUMGO-8X      (f)UBRAB-8X</div>

<div align="center">Fig. 16 Procedures render based on gpt-4o</div>

## A3.4 procedure render in ablation experiment

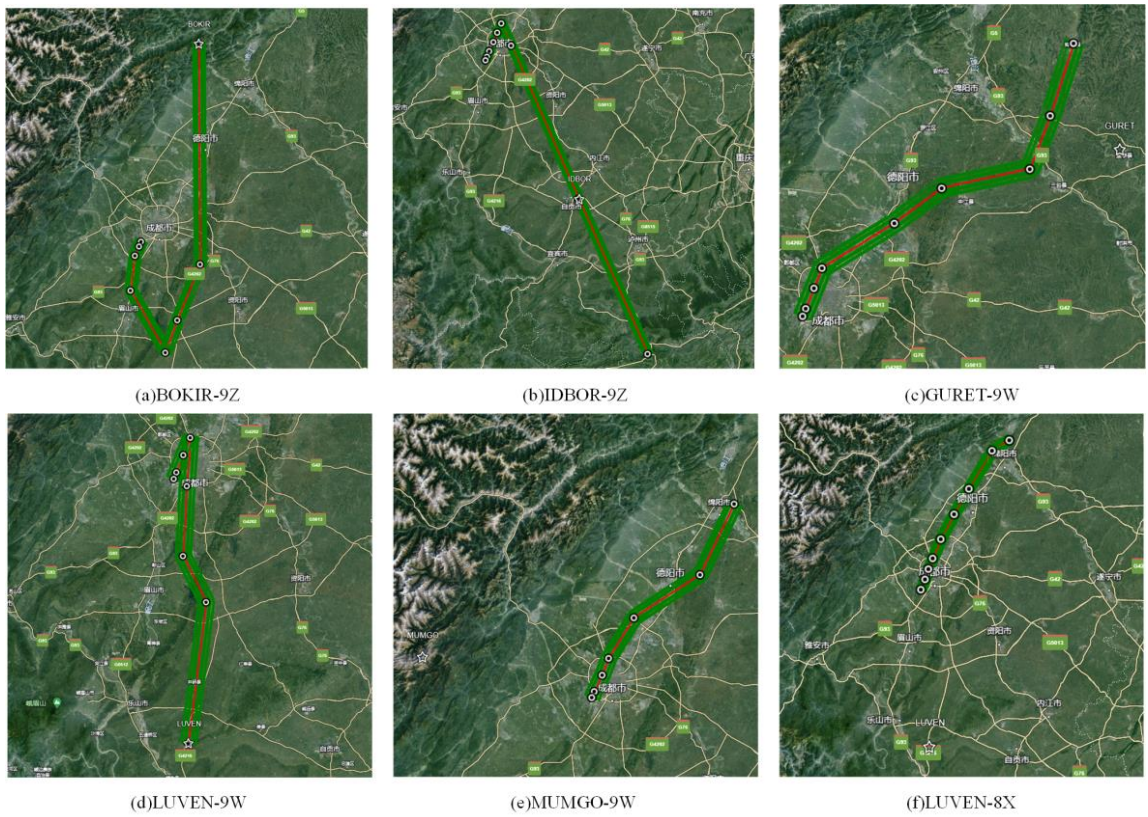

<div align="center">(a)BOKIR-9Z      (b)IDBOR-9Z      (c)GURET-9W</div>

<div align="center">(d)LUVEN-9W      (e)MUMGO-9W      (f)LUVEN-8X</div>

(g)MUMGO-8X

(h)UBRAB-8X

Fig. 17 Procedures render in ID1

(a)BOKIR-9Z

(b)IDBOR-9Z

(c)GURET-9W

(d)LUVEN-9W

(e)MUMGO-9W

(f)LUVEN-8X

(g)MUMGO-8X

(h)UBRAB-8X

Fig. 18 Procedures render in ID2

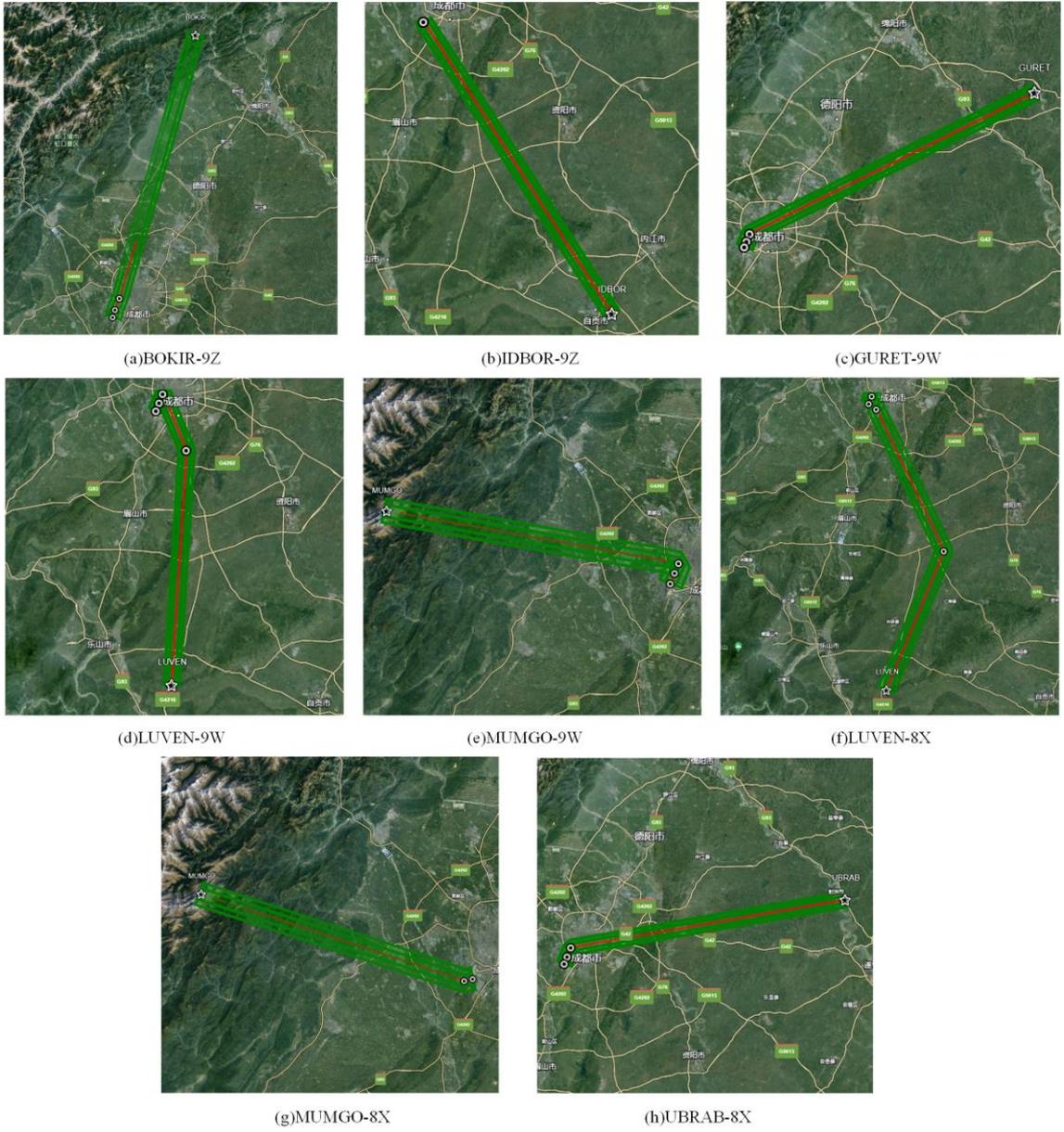

(a)BOKIR-9Z  (b)IDBOR-9Z  (c)GURET-9W

(d)LUVEN-9W  (e)MUMGO-9W  (f)LUVEN-8X

(g)MUMGO-8X  (h)UBRAB-8X

Fig. 19 Procedures render in ID3

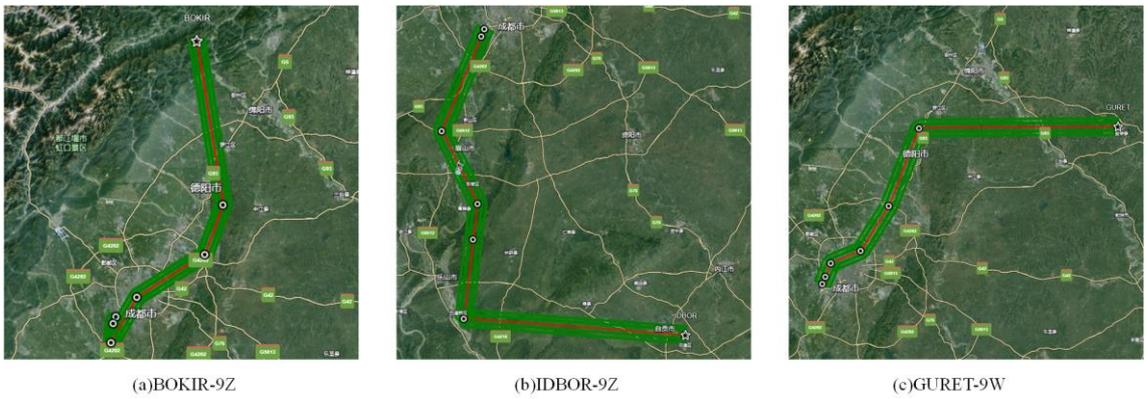

(a)BOKIR-9Z  (b)IDBOR-9Z  (c)GURET-9W

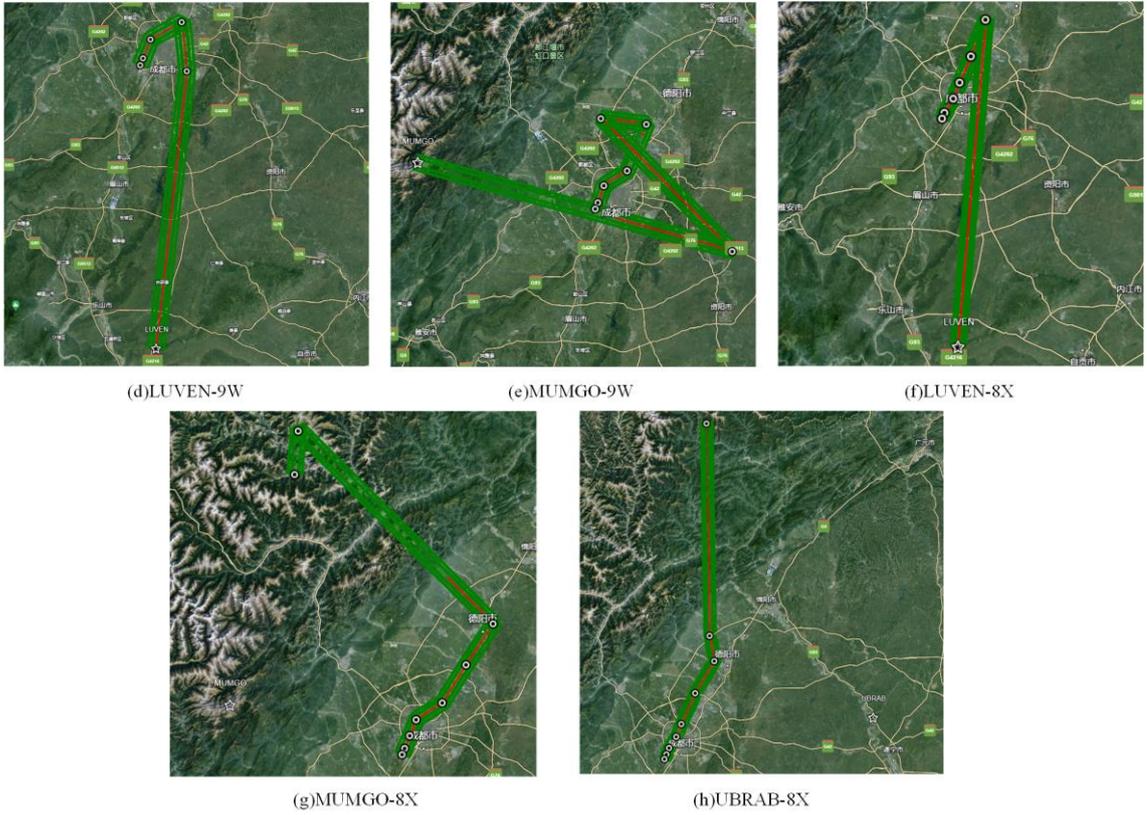

Fig. 20 Procedures render in ID4

## A3.5 procedure render in human involved experiment

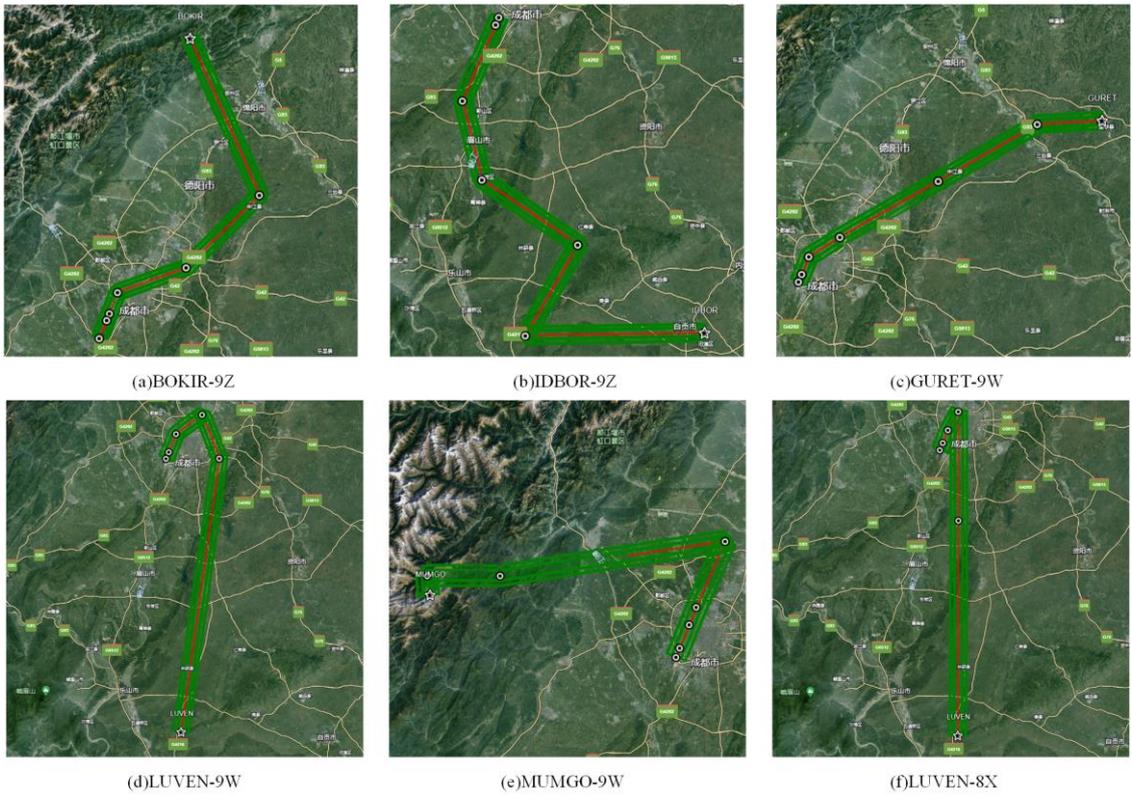

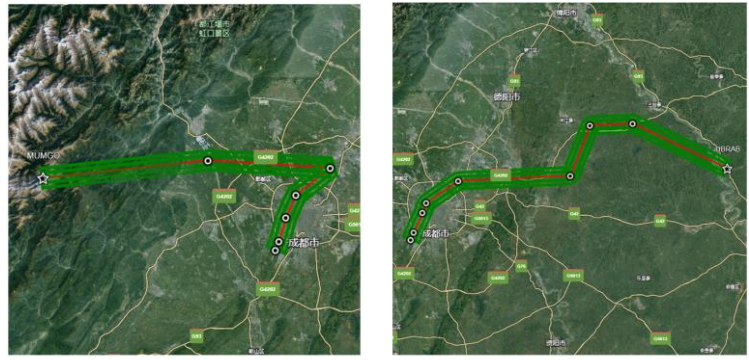

(g)MUMGO-8X　　　　　　　　　　　(h)UBRAB-8X

Fig.21 Procedures render with human involved

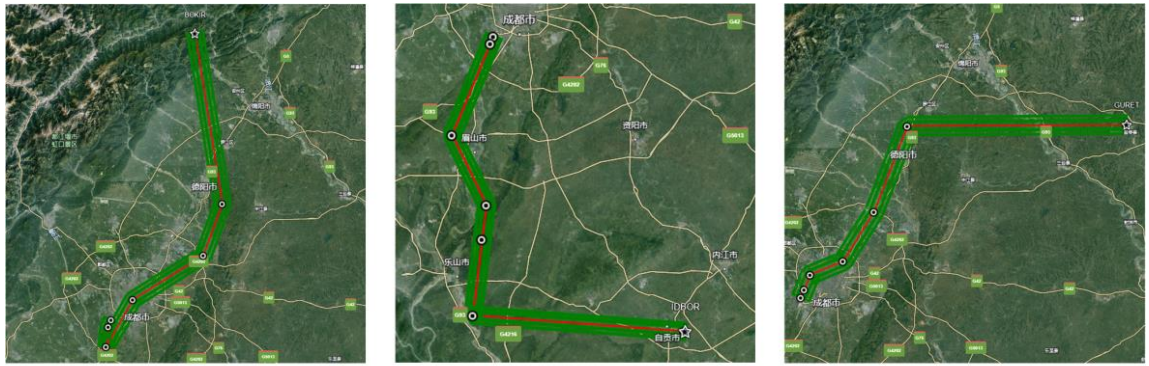

(a)BOKIR-9Z　　　　　　　(b)IDBOR-9Z　　　　　　　(c)GURET-9W

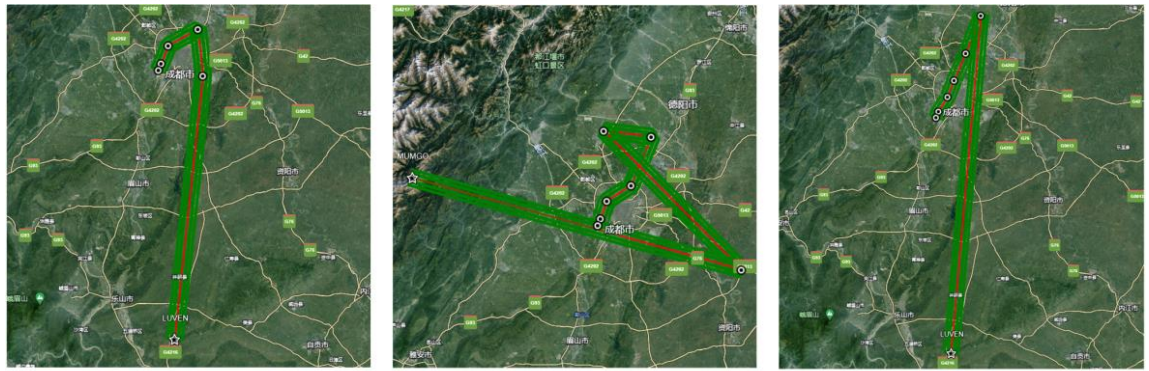

(d)LUVEN-9W　　　　　　　(e)MUMGO-9W　　　　　　　(f)LUVEN-8X

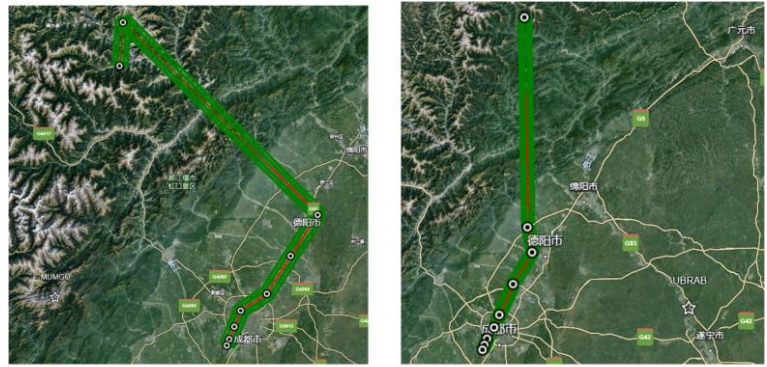

(g)MUMGO-8X　　　　　　　　　　　(h)UBRAB-8X

Fig.22 Procedures render without human involved